\documentclass[sigconf, nonacm]{acmart}
\usepackage{algorithm}
\usepackage{algpseudocode}
\usepackage{amsmath}
\usepackage{graphicx}
\usepackage{subcaption}
\usepackage{multirow} 

\begin{document}
\title{An Efficient Gradient-Aware Error-Bounded Lossy Compressor for Federated Learning}


\author{Zhijing Ye}
\affiliation{%
  \institution{Stevens Institute of Technology}
  \city{Hoboken}
  \state{NJ, USA}
}
\email{zye25@stevens.edu}

\author{Sheng Di}
\affiliation{%
  \institution{Argonne National Laboratory}
  \city{Lemont}
  \state{IL, USA}
}
\email{sdi1@anl.gov}

\author{Jiamin Wang}
\affiliation{%
  \institution{Stevens Institute of Technology}
  \city{Hoboken}
  \state{NJ, USA}
}
\email{jwang259@stevens.edu}

\author{Zhiqing Zhong}
\affiliation{%
  \institution{Stevens Institute of Technology}
  \city{Hoboken}
  \state{NJ, USA}
}
\email{zzhong9@stevens.edu}

\author{Zhaorui Zhang}
\affiliation{
  \institution{Hong Kong Polytechnic University}
  \city{Hung Hom}
  \state{Hong Kong}
}
\email{zhaorui.zhang@polyu.edu.hk}

\author{Xiaodong Yu}
\affiliation{%
  \institution{Stevens Institute of Technology}
  \city{Hoboken}
  \state{NJ, USA}
}
\email{xyu38@stevens.edu}

\begin{abstract}
Federated learning (FL) enables collaborative model training without exposing clients’ private data, but its deployment is often constrained by the communication cost of transmitting gradients between clients and the central server, especially under system heterogeneity where low-bandwidth clients bottleneck overall performance. Lossy compression of gradient data can mitigate this overhead, and error-bounded lossy compression (EBLC) is particularly appealing for its fine-grained utility–compression tradeoff. However, existing EBLC methods (e.g., SZ), originally designed for smooth scientific data with strong spatial locality, rely on generic predictors such as Lorenzo and interpolation for entropy reduction to improve compression ratio. Gradient tensors, in contrast, exhibit low smoothness and weak spatial correlation, rendering these predictors ineffective and leading to poor compression ratios.


To address this limitation, we propose an EBLC framework tailored for FL gradient data to achieve high compression ratios while preserving model accuracy. The core of it is an innovative prediction mechanism that exploits temporal correlations across FL training rounds and structural regularities within convolutional kernels to reduce residual entropy. The predictor is compatible with standard quantizers and entropy coders and comprises (1) a cross-round magnitude predictor based on a normalized exponential moving average, and (2) a sign predictor that leverages gradient oscillation and kernel-level sign consistency. Experiments show that this new EBLC yields up to 1.53× higher compression ratios than SZ3 with lower accuracy loss. Integrated into a real-world FL framework, APPFL, it reduces end-to-end communication time by 76.1\%–96.2\% under various constrained-bandwidth scenarios, demonstrating strong scalability for real-world FL deployments.

\end{abstract}

\maketitle



\section{Introduction}






Federated learning (FL) is a distributed and privacy-preserving training paradigm that enables multiple clients to collaboratively train a shared global model while keeping their local data private. FL holds strong potential in application domains such as autonomous driving and medical AI, where data from multiple sources are required to train robust models, but direct data sharing is restricted by privacy regulations~\cite{rieke2020future, pokhrel2020federated}. In the medical domain, a representative use case is the collaborative training of convolutional neural networks (CNNs) for magnetic resonance imaging (MRI) analysis, such as detecting tumors or segmenting anatomical structures~\cite{darestani2025convolutional}. Federated learning enables multiple hospitals to jointly train such models without exposing sensitive patient data~\cite{guo2021multi}.

Despite its promise, communication overhead remains a critical bottleneck in FL~\cite{yang2022flash,bonawitz2019towards,gabrielli2023survey}. In each round of training, participating clients must transmit model updates to the central server. As model sizes and the number of clients increase, this communication cost often dominates the end-to-end training time. The challenge is further exacerbated in real-world deployments, where clients are heterogeneous and network bandwidth is limited. A client on a 4G-LTE connection typically sustains only 20–40 Mbps, while Wi-Fi users may reach 100–200 Mbps, and those with fiber broadband can exceed 1 Gbps. Such disparity can introduce up to a 50$\times$ difference in upload latency. Consequently, the overall training round time is constrained by the slowest participant, severely degrading throughput~\cite{kairouz2021advances,bonawitz2019towards}. Therefore, to ensure efficient and robust FL, reducing communication cost is essential.

A promising technique to address this challenge is data reduction, which can shrink the overall communication cost.  Existing approaches can be broadly categorized into sparsification and quantization. \textit{Sparsification} methods (e.g., TopK~\cite{aji2017sparse}) achieve high compression ratios by transmitting only a small subset of gradients while setting the rest to zero. These methods are lightweight and easy to deploy, but they essentially discard the majority of the update information. As a result, the induced information loss can accumulate and in some cases lead to significant accuracy degradation. \textit{Quantization} methods, on the other hand, retain all coordinates but represent them with fewer bits (e.g., QSGD~\cite{alistarh2017qsgd}), thereby reducing communication volume while preserving more information than sparsification. Within this family, error-bounded lossy compressors (EBLCs)~\cite{lindstrom2014fixed,sz3,ainsworth2019multilevel,sz2} provide a more principled approach by explicitly constraining the compression error within a user-specified bound. This fine-grained error control allows EBLCs to maintain model accuracy while still achieving potentially high compression ratios, making them a strong candidate solution to FL’s communication bottlenecks.

A handful of previous works~\cite{wilkins2024fedsz, zhang2025fedcspc, zhang2025fedefsz} have directly applied existing EBLCs to FL gradient data. However, their compression performances is typically suboptimal because most EBLCs, such as SZ3~\cite{sz3}, were originally designed for scientific datasets characterized by high smoothness and strong spatial locality. Most existing EBLCs incorporate crucial prediction components implemented using generic algorithms such as Lorenzo and interpolation, which are intended to exploit data regularities to produce accurate local estimates. Accurate prediction yields concentrated, low-entropy residuals for encoder, enabling significantly higher compression ratios~\cite{tao2017significantly}. While these predictors work well for structured scientific data, they fail to perform effectively on gradient tensors, whose spatial patterns are highly irregular and stochastic. This mismatch between predictor design and the inherent characteristics of gradient distributions fundamentally limits the achievable compression ratio. Therefore, it is essential to develop a novel prediction mechanism tailored to the properties of gradient data to enhance the effectiveness of EBLCs in FL.


In this work, we propose an EBLC framework tailored for FL gradients. The framework introduces a new predictor specifically optimized for gradient data that can seamlessly integrate with standard quantizer, entropy coder, and lossless compression modules used in state-of-the-art EBLC. Our new design can achieve significantly higher compression ratios than existing EBLCs while maintaining the same level of error control. The key innovation lies in exploiting two persistent regularities in gradients: (i) temporal correlation across rounds, captured with an EMA-based magnitude predictor and oscillation-aware sign hints; and (ii) intra-kernel sign consistency, exploited through a kernel-level sign predictor that assigns dominant signs to predictable kernels and encodes both sign and positions using a two-level bitmaps. By jointly leveraging these temporal and spatial regularities, our compressor produces more compact residuals that are efficiently encoded under the same error bound. This dual-domain redundancy elimination substantially reduces communication cost without undermining convergence or final accuracy. Combined with its low overhead and drop-in compatibility, it enables EBLCs to become a practical and scalable solution for real-world FL deployments.

We integrate our proposed EBLC into APPFL, an open-source production-grade FL framework, and conduct comprehensive evaluations on widely used CNN models (ResNet-18/34, Inception V1/V3) across standard datasets (Fashion-MNIST, CIFAR-10, Caltech101). We compare against SZ3, a state-of-the-art, highly optimized EBLC widely considered to be near the practical limit of error-bounded compression, and QSGD, a popular gradient compression method. The results show that our proposed EBLC achieves up to 1.53$\times$ and 2.11$\times$ higher compression ratios than SZ3 and QSGD, respectively, with lower accuracy degradation. It consequently reduces end-to-end communication time by 76.1\%–96.2\% under constrained-bandwidth settings. In summary, this paper makes the following contributions:
\vspace{-1ex}
\begin{itemize}
    \item We propose an EBLC framework specifically designed to achieve ultra-high compression ratios for FL gradient data, effectively mitigating communication bottlenecks while preserving model accuracy.
    \item We design an innovative predictor that exploits temporal correlations across FL training rounds and structural regularities within convolutional kernels, substantially reducing residual entropy to enable higher compression ratios.
    \item We integrate it into APPFL and conduct comprehensive evaluations. Results show that our method consistently achieves higher compression ratios than SZ3 (up to 1.53×) with maintained training accuracy, and reduces end-to-end communication time by 76.1\%–96.2\%.
    
\end{itemize}

The rest of this paper is organized as follows. Section~2 provides background, and Section~3 motivates our design. Section~4 presents the proposed compressor, and Section~5 reports evaluation results. Section~6 discusses limitations and future work, Section~7 reviews related work, and Section~8 concludes.


\section{Background}
This section provides background on compression in federated learning and error-bounded lossy compressors. We first discuss how data reduction mitigates communication overhead in FL and then summarize the typical four-stage EBLC pipeline.

\subsection{Compression for Communication Efficiency}
In federated learning (FL), communication occurs in each training round when clients send gradients to the parameter server. Let $S$ denote the size of the transmitted update and $B$ the available bandwidth. The original communication time without compression can be approximated as $T_{\text{ori}} = \frac{S}{B}$.
\begin{figure}
    \centering
    \includegraphics[width=0.8\linewidth]{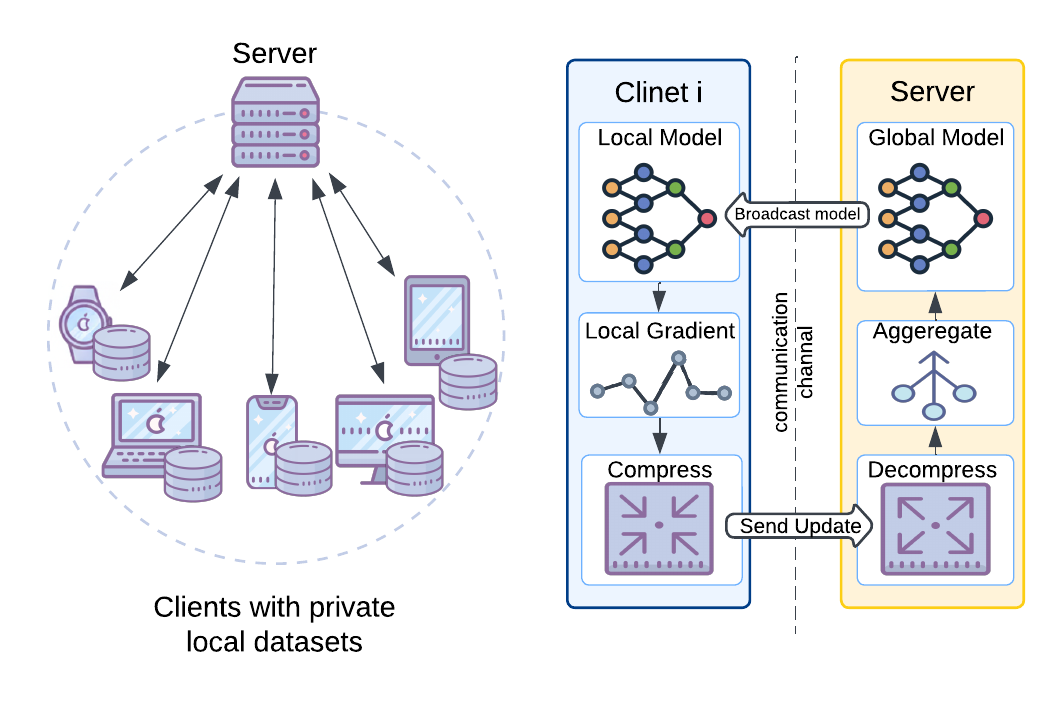}
    \caption{Illustration of communication-efficient federated learning with compression.}
    \vspace{-0.5em}
    \label{fig:compression in fl}
\end{figure}

The typical workflow of adopting data reduction techniques to reduce communication cost in FL is described in Figure ~\ref{fig:compression in fl}. First of all, each client performs a local training, obtaining its local gradient and applies a compression algorithm to reduce its size. The compressed gradient, with size $S'$, is then transmitted to the parameter server. Upon reception, the server first decompresses to recover the original gradients, and then aggregate them to update the global model.

To evaluate how data reduction influences communication efficiency, we establish a quantitative model for the end-to-end communication time. The total communication time with data reduction $T_{\text{comm}}$ is expressed as:
\begin{equation}
    T_{\text{comm}} = T_{\text{comp}} + \frac{S'}{B} + T_{\text{decomp}}
    \label{eq:comm time}
\end{equation}

where $T_{\text{comp}}$ and $T_{\text{decomp}}$ denote the compression and decompression time. To better understand the performance benefit of compression, we analyze the ratio between compressed and uncompressed communication time:
\begin{equation}
    \frac{T_{\text{comm}}}{T_{\text{ori}}} = \frac{T_{\text{comp}} + T_{\text{decomp}}}{S/B} + \frac{1}{CR}
    \label{eq:time ratio}
\end{equation}
where $CR = S/S'$ is the compression ratio. The first term quantifies the relative compression and decompression overhead with respect to the original communication time, which is typically small under bandwidth-limited environments. The second term directly captures the benefit of data reduction, showing that a higher $CR$ leads to greater savings in communication time. Therefore, the overall benefit of data reduction is strongly tied to achieving higher $CR$.

\subsection{Error-Bounded Lossy Compressors}
EBLCs guarantee that compression errors stay within a user-specified tolerance while achieving substantial data reduction. Among them, the SZ-family compressors have emerged as a widely adopted state-of-the-art solution in scientific computing.

\begin{figure}[t]
    \centering
    \includegraphics[width=0.8\linewidth]{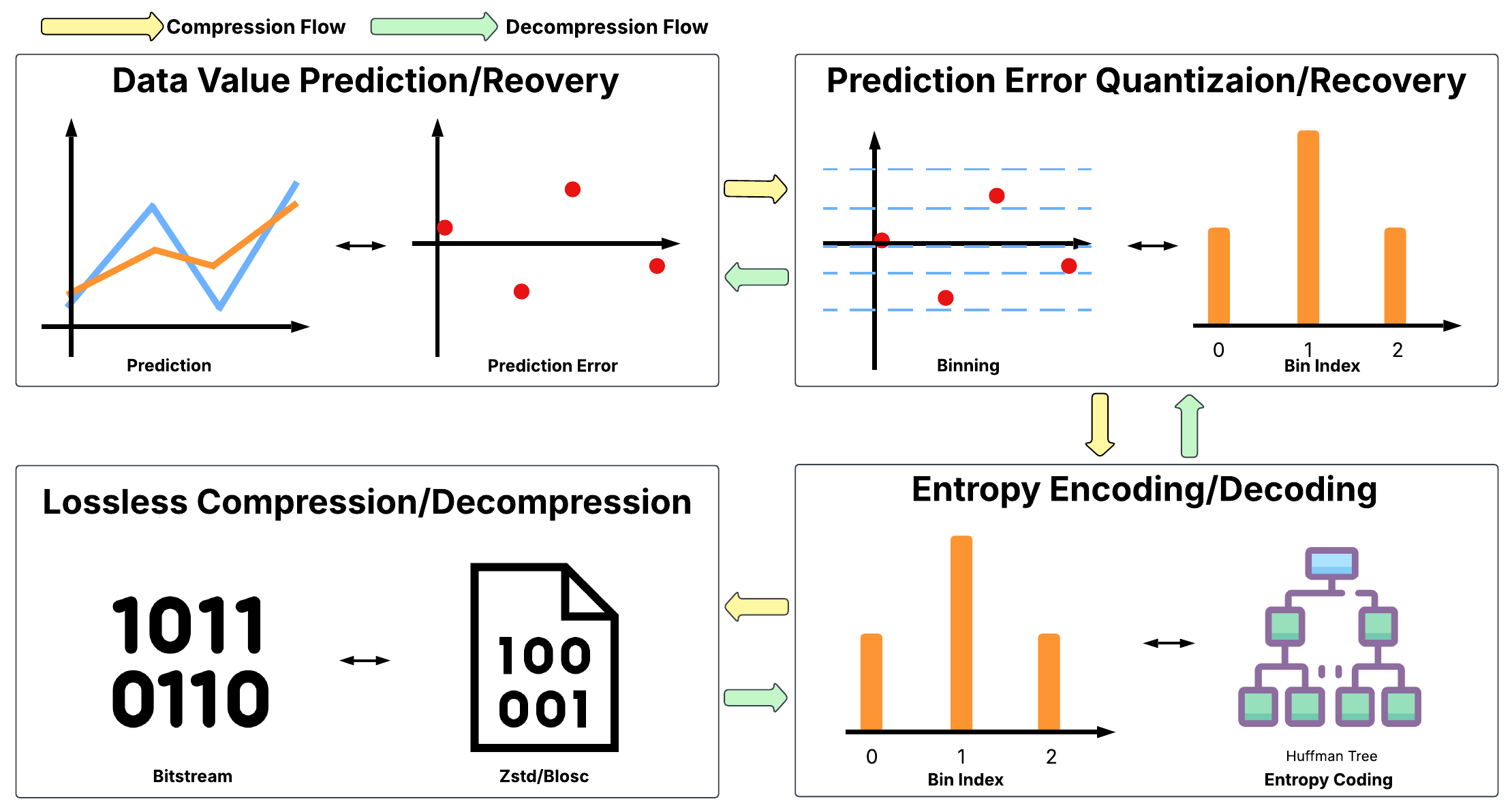}
    \caption{Typical four-stage pipeline for SZ compressor includes prediction, quantization, entropy encoding and lossless compression.}
    \label{fig:EBLC Pipeline}
    \vspace{-1em}
\end{figure}

As shown in Figure~\ref{fig:EBLC Pipeline}, SZ follows a four-stage pipeline:

\begin{itemize}
    \item \textbf{Stage 1: Prediction.}
    Each data point is estimated from its local context (e.g., neighboring or previously reconstructed samples), and only the residual is stored. This step concentrates residuals around zero, lowering entropy and enabling effective compression.
    
    \item \textbf{Stage 2: Quantization.}
    Residuals are discretized into bins defined by a user-specified error bound, ensuring the reconstruction error remains within tolerance. This discretization increases symbol frequency and further reduces entropy.
    
    \item \textbf{Stage 3: Encoding.}
    Quantized residuals are entropy coded (e.g., via Huffman), assigning shorter codes to frequent values and longer codes to rare ones to minimize average bit length.
    
    \item \textbf{Stage 4: Lossless compression.}
    A general-purpose compressor (e.g., Zstd or Blosc) is applied to the coded stream to remove remaining redundancy and improve the final compression ratio.
\end{itemize}

Building on this general framework, successive versions of the SZ compressor family have introduced increasingly sophisticated prediction models and optimization techniques to enhance accuracy and compression efficiency.



\textbf{SZ3}~\cite{sz3,SZ3_BD} represents a major advancement in the SZ family, introducing a dynamic spline interpolation predictor that replaces linear regression to reduce metadata and improve prediction accuracy. The resulting low-entropy residuals achieve higher compression ratios, particularly under large error bounds. \textbf{cuSZP}~\cite{huang2023cuszp} extends SZ3 to GPUs with an optimized prediction–quantization–encoding pipeline. It employs a parallel Lorenzo predictor with predictor fusion and pipelined quantization, leveraging GPU memory hierarchy and asynchronous execution to deliver high compression ratios and orders-of-magnitude speedups over CPU-based compressors.

 These compressors are highly effective for scientific datasets, where smoothness and local continuity lead to small, low-entropy residuals that compress extremely well. However, gradient data in federated learning typically lacks such smoothness, making generic predictors ineffective and leaving substantial room for improvement.

\section{Limitations in Existing EBLC}
This section analyzes two major limitations of existing EBLCs in the context of FL gradient data: the ineffectiveness of their generic predictors and the overlooked temporal patterns in gradients.

\subsection{Generic Predictors are Ineffective}

Generic predictors in EBLCs fail to perform well on gradient data. Compressors such as SZ employ generic predictors like the Lorenzo predictor and interpolation, which are designed to exploit smoothness and local continuity in scientific datasets. In the one-dimensional case, the Lorenzo predictor estimates the current value as its previous neighbor. The interpolation predictor, by contrast, uses adjacent known neighbors to predict the current point as their linear combination.

\begin{figure}
    \centering
    \includegraphics[width=\linewidth]{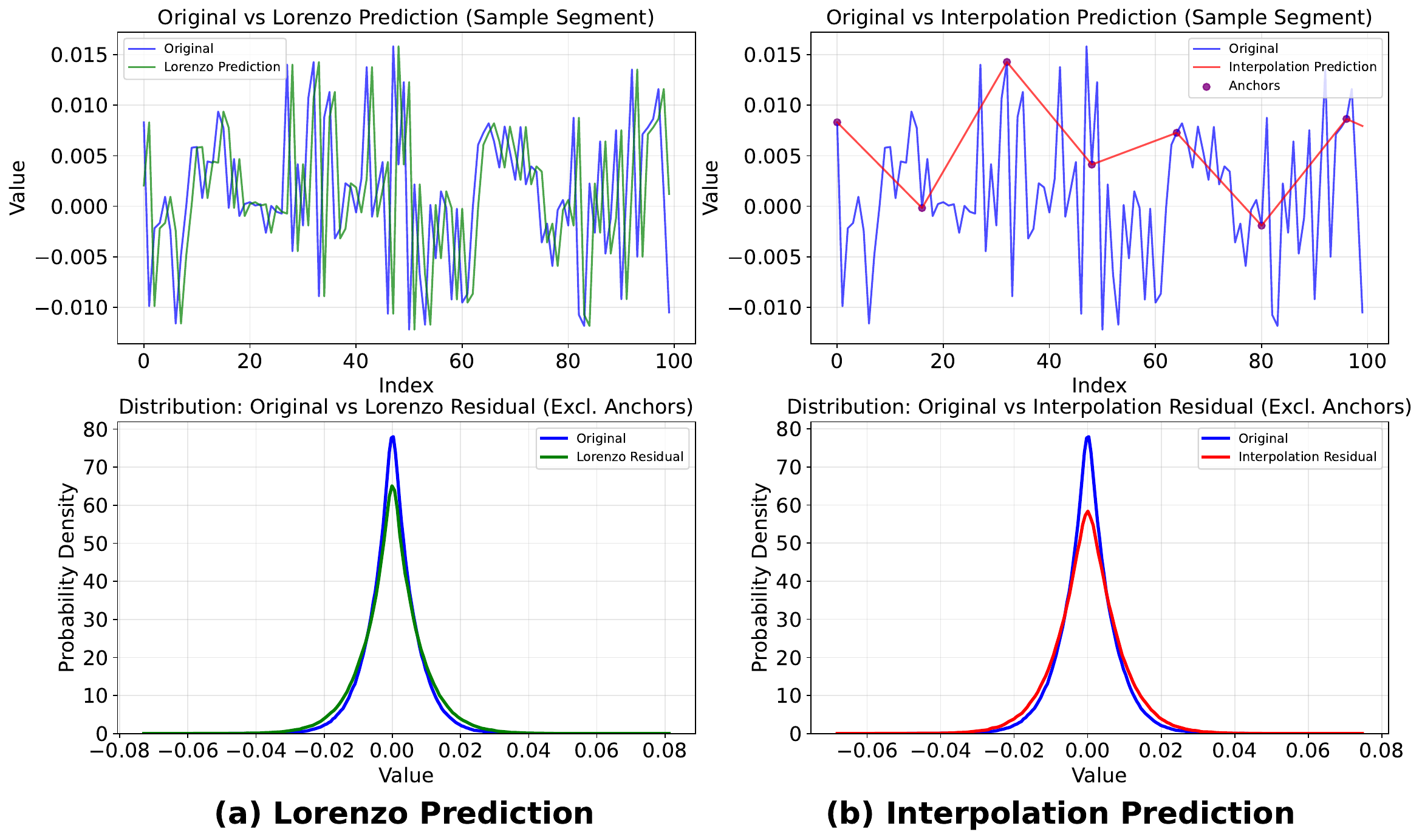}
    \caption{The sample prediction result of (a) Lorenzo and (b) Interpolation on real gradient data and the comparison of the original gradient and the residual after prediction.}
    \label{fig:Generic Predictor}
    \vspace{-1em}
\end{figure}

Figure~\ref{fig:Generic Predictor} demonstrates Lorenzo and interpolation predictor on real gradient data. As shown, both predictors fail to follow the noisy and irregular patterns of gradients. The predicted values deviate significantly from the originals, leading to residuals that are neither centered tightly around zero nor lower in entropy compared to the raw data. In fact, as illustrated by the residual distributions, prediction sometimes makes the data harder to compress, since the residual variance may be larger than that of the original values.

These observations highlight a fundamental mismatch: unlike scientific datasets, gradient data in federated learning does not exhibit smoothness or local continuity. Consequently, the assumptions underlying Lorenzo and interpolation predictors often do not hold, limiting their effectiveness. To design an efficient compressor for FL, we must therefore move beyond generic predictors and investigate what unique properties gradient data actually possesses and how they can be exploited.

\subsection{Grad Temporal Patterns are Unleveraged}
\label{subsec:temporal_patterns}
\begin{figure}[b]
    \centering
    \begin{subfigure}{0.48\linewidth}
        \centering
        \captionsetup{font=scriptsize}
        \includegraphics[width=\linewidth]{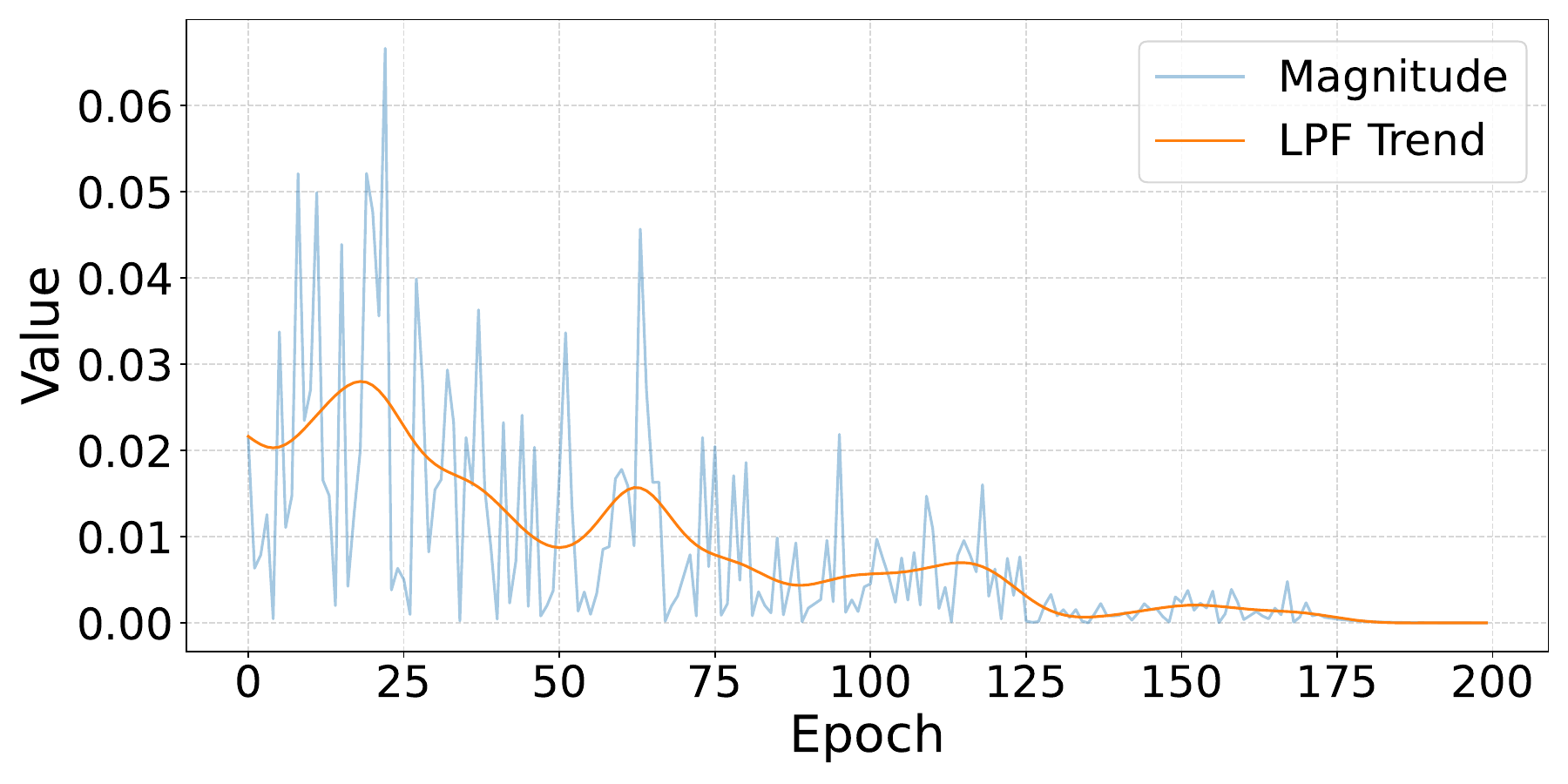}
        \caption{Gradient Magnitude Trend}
        \label{fig:sub1}
    \end{subfigure}
    \hfill
    \begin{subfigure}{0.48\linewidth}
        \centering
        \captionsetup{font=scriptsize}
        \includegraphics[width=\linewidth]{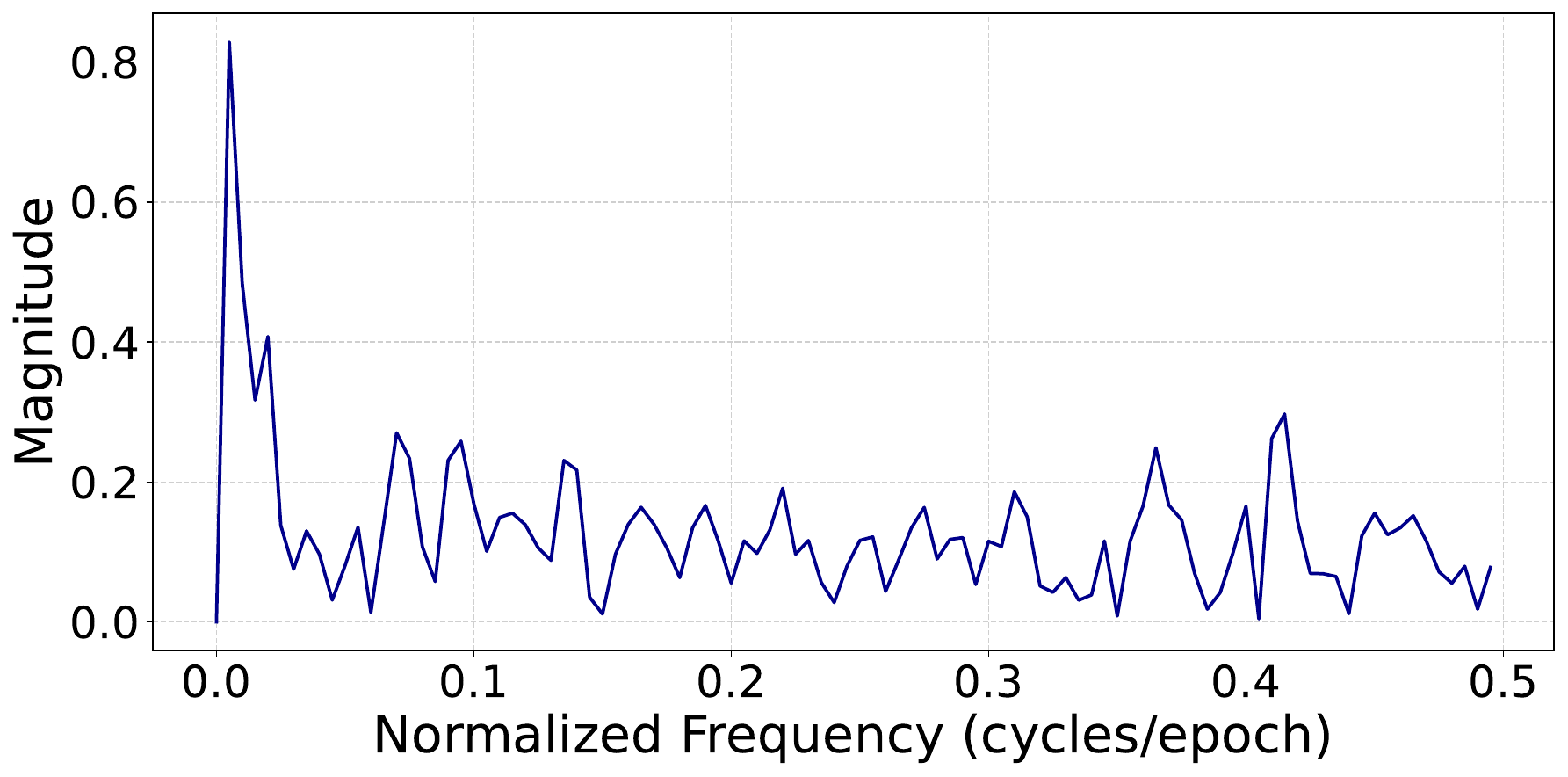}
        \caption{Magnitude Spectrum}
        \label{fig:sub2}
    \end{subfigure}

    \caption{An example gradient magnitude trend across 200 epochs and corresponding magnitude spectrum.}
    \label{fig:envelope and spectrum}
\end{figure}

Although raw gradients are difficult to predict spatially, their iterative generation introduces temporal dependencies that give rise to exploitable patterns. In optimization, each gradient is computed based on the model parameters and loss from the previous step, creating sequential correlations over time~\cite{wang2019adaptive}. By analyzing their evolution across training rounds, we find that both the magnitudes and the signs of gradients exhibit meaningful temporal regularities that can inform prediction.

Figure~\ref{fig:envelope and spectrum} (a) shows gradient magnitudes and the trend after applying a low-pass filter. Figure~\ref{fig:envelope and spectrum} (b) shows its frequency spectrum across training epochs. We can see that the magnitudes generally decrease as training progresses, and their variation is dominated by low-frequency components, while high-frequency noise accounts for only a smaller portion. This indicates that, although raw gradients themselves are difficult to predict, their magnitudes follow a clear downward trend with low-frequency dynamics that can be exploited.

In addition, prior work ~\cite{grad_oscillation} has identified oscillatory behavior in gradient directions under full-batch gradient descent. In this setting, the update rule can be written as:
\begin{equation}
    \text{(GD):} \quad 
    \theta_{k+1} = \theta_k - \eta \nabla \mathcal{L}_n(\theta_k),
    \label{eq:GD}
\end{equation}
where $\theta_k$ denotes the model parameters at iteration $k$, $\eta$ is the learning rate, and $\nabla \mathcal{L}_n(\theta_k)$ is the gradient of the empirical loss over all $n$ samples. 

\begin{figure}[t]
    \centering
    \includegraphics[width=\linewidth]{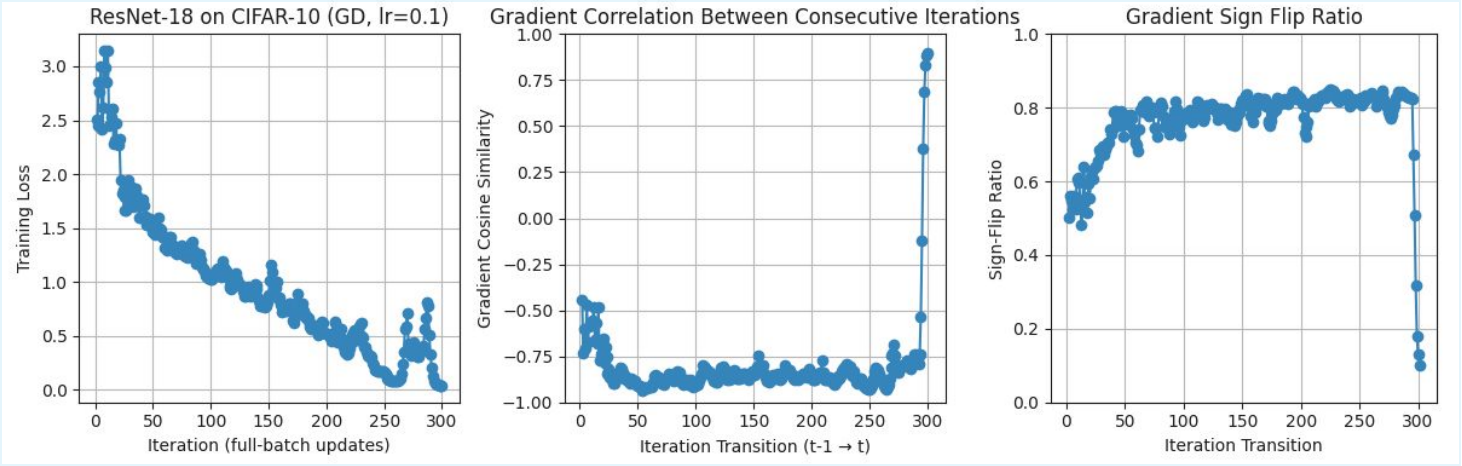}
    \caption{Gradient oscillation across epochs under full-batch gradient descent.}
    \label{fig:gradient oscillation}
\end{figure}

The gradient correlation is used to quantify the directional correlation. The gradient correlation between arbitrary iterations $j$ and $k$ can be expressed as:
\begin{equation}
    \text{(Gradient correlation):} \quad
    \mu(j,k) := \frac{\langle \nabla \mathcal{L}_n(\theta_j), 
    \nabla \mathcal{L}_n(\theta_k) \rangle}
    {\|\nabla \mathcal{L}_n(\theta_j)\| \cdot 
    \|\nabla \mathcal{L}_n(\theta_k)\|}
    \label{eq: gradient correlation}
\end{equation}
Empirical studies show strong correlation or anti-correlation between successive gradients as illustrated in Figure~\ref{fig:gradient oscillation}. Consequently, the sign of a gradient can often be predicted with relatively high probability from its historical behavior.

Together, these findings reveal that while raw gradients are difficult to predict directly in space, their magnitudes and signs contain consistent temporal patterns. Leveraging these patterns enables more accurate gradient prediction and, in turn, more efficient gradient compression. Although prior works~\cite{mishchenko2024distributed, gorbunov2021marina} have also leveraged temporal relationships among gradients, they primarily focus on sparsification-based approaches. To the best of our knowledge, none has explicitly exploited temporal magnitude and sign patterns from the perspective of error-bounded lossy compression.

\begin{figure*}[t]
    \centering
    \includegraphics[width=0.95\linewidth]{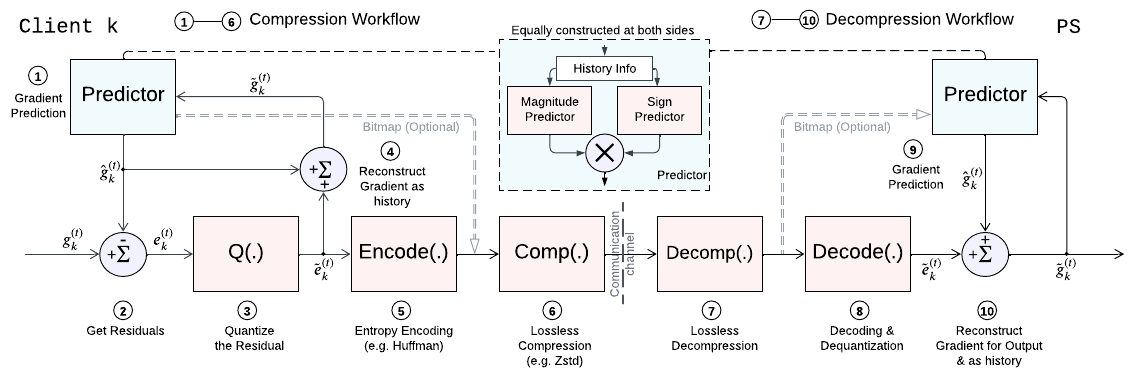}
    \vspace{-1.5em}
    \caption{Overview of our proposed high-ratio error-bounded gradient compression framework.}
    \label{fig:overview of proposed compressor}
\end{figure*}

\section{Framework Design}
In this section, we present the design of our work, an efficient gradient-aware EBLC tailored for FL. The core idea of our design is a novel prediction mechanism that leverages both temporal correlations across training rounds and structural properties within convolutional layers. By producing residuals with lower entropy, this predictor enables significant improvement in compression ratio without compromising error control.

\subsection{Overview of the Framework}
The proposed compressor follows the standard four-stage architecture of EBLCs, consisting of prediction, quantization, entropy coding, and lossless compression, while enhancing the prediction stage to better accommodate the characteristics of gradient data, as illustrated in Figure~\ref{fig:overview of proposed compressor}. The design is compatible with most quantizers and entropy coders used in EBLCs, allowing flexible configuration for different application scenarios and requirements.

In the compression workflow, each client performs compression locally. After completing local training, client $k$ obtains its gradient $g^{(t)}_k$ at epoch $t$ and feeds into the compressor. \textcircled{1} The predictor then estimates $\hat g^{(t)}_k$ by leveraging historical reconstructed gradients
\[
\hat g_t = f\bigl(\tilde g_{t-1},\,\tilde g_{t-2},\dots,\,\tilde g_{t-p}\bigr),
\]
where $f(\cdot)$ is our online predictor that integrates both magnitude and sign information. \textcircled{2}The residual is computed as $e^{(t)}_k = g^{(t)}_k - \hat g^{(t)}_k$, which removes the predictable low-frequency component of the gradient. Because this residual is highly concentrated around zero, its entropy is greatly reduced, making it more suitable for compression. \textcircled{3}The residual is then quantized under a user-specified error bound $\Delta$:
\[
\tilde e^{(t)}_k = Q_\Delta(e^{(t)}_k), \quad
\bigl|\tilde e^{(t)}_k - e^{(t)}_k\bigr| \le \Delta,  
\]
where $Q_\Delta(\cdot)$ denotes an error-bounded quantizer with user-defined error bound $\Delta$. Only the quantized residual $Q_\Delta(e^{(t)}_k)$ is transmitted, \textcircled{4} while the reconstructed gradient $\tilde g^{(t)}_k = \hat g^{(t)}_k + \tilde e^{(t)}_k$ is stored locally as history for next prediction. \textcircled{5}The quantized residuals are then entropy encoded (e.g., Huffman coding) and \textcircled{6}further compressed by a lightweight lossless compressor such as Zstd or Blosc, yielding a compact binary stream sent to the parameter server.

In the decompression workflow,the parameter server performs decompression symmetrically to reconstruct the transmitted gradients. \textcircled{7}The received payload is first performed lossless decompressed to reconstruct the bit stream. \textcircled{8}Then it is decoded and dequantized to recover the quantized residual $\tilde e^{(t)}_k$. \textcircled{9}The predictor on the PS reproduces the gradient prediction $\hat g^{(t)}_k$. \textcircled{10}This prediction is then combined with the quantized residual $\tilde e^{(t)}_k$ to reconstruct the gradient
$\tilde g^{(t)}_k$. Importantly, the predictors on the client and server are identically constructed and operate in exactly the same way. Both sides update their predictors using the reconstructed gradients $\tilde g^{(t)}_k$, ensuring that the prediction states remain synchronized without requiring any additional communication. The only exception arises in the sign-prediction stage, where a small two-level bitmap may be transmitted to handle cases of kernel-level dominant sign prediction, which we will discuss in Section ~\ref{subsection: sign prediction}. However, this overhead is negligible compared to the overall payload size. As a result, beyond the compressed residuals, no extra metadata needs to be exchanged between the client and server.

\subsection{Gradient Magnitude Predictor}
Exploiting the smoother low-frequency trends of gradient magnitudes across epochs, we design a lightweight online predictor that applies per-epoch normalization and an exponential moving average (EMA) to adapt to the non-stationary dynamics of federated learning.

We empirically compare this design against several alternatives, including Lorenzo predictors, moving averages with sliding window sizes of 3 and 5, First-order autoregressive (AR) model, and EMA without normalization. As shown in Table~\ref{tab:mag_pred_ablation_t}, the combination of EMA and normalization consistently achieves better prediction accuracy while remaining lightweight. We also explored heavier alternatives such as Kalman filters~\cite{kalman1960new} and Transformer-based time-series models~\cite{zhou2021informer,liu2023itransformer, wu2021autoformer}, but they provided no notable improvement under FL conditions and incurred substantial computational and memory overhead. Overall, EMA with normalization offers the best trade-off between accuracy, robustness, and efficiency, making it a practical choice for improving gradient compressibility in federated learning.

\begin{table}
\centering
\caption{Ablation on gradient magnitude predictors. Lower is better for MSE; higher is better for Corr.}
\label{tab:mag_pred_ablation_t}
\resizebox{\linewidth}{!}{
\setlength{\tabcolsep}{3pt}
\begin{tabular}{lcccccc}
\toprule
 & Lorenzo & MA ($w{=}3$) & MA ($w{=}5$) & AR(1) & EMA (No Norm) & \textbf{EMA (Norm)} \\
\midrule
MSE  & $2.50\times 10^{-4}$ & $1.90\times 10^{-4}$ & $1.89\times 10^{-4}$ & $1.90\times 10^{-4}$ & $1.38\times 10^{-4}$ & $\mathbf{9.18\times 10^{-5}}$ \\
Corr & 0.3196 & 0.3701 & 0.3887 & 0.3196 & 0.4360 & \textbf{0.5608} \\
\bottomrule
\end{tabular}
}
\end{table}

\begin{algorithm}
\footnotesize
\caption{Gradient Magnitude Predictor (client $k$)}
\label{alg:abs_value_predictor}
\begin{algorithmic}[1]
\Statex \textbf{Input:} Previous reconstructed absolute gradient $\tilde{a}^{(t-1)}_k$, current absolute gradient mean $\mu_{\text{curr}}$ and standard deviation $\sigma_{\text{curr}}$, previous memory $m$, decay factor $\beta$
\Statex \textbf{Output:} Predicted magnitude $\widehat{a}^{(t)}_k$, updated memory $m$
\State $\mu_{\text{prev}} \gets \operatorname{mean}(\tilde{a}^{(t-1)}_k)$;\quad $\sigma_{\text{prev}} \gets \operatorname{std}(\tilde{a}^{(t-1)}_k)$
\State $z_{\text{prev}} \gets \dfrac{\tilde{a}^{(t-1)}_k - \mu_{\text{prev}}}{\sigma_{\text{prev}}}$
\State $z_{\text{pred}} \gets (1-\beta)\cdot m + \beta\cdot z_{\text{prev}}$ \Comment{prediction in normalized space}
\State $\widehat{a}^{(t)}_k \gets z_{\text{pred}} \cdot \sigma_{\text{curr}} + \mu_{\text{curr}}$ \Comment{de-normalize to current scale}
\State $m \gets z_{\text{pred}}$ \Comment{update predictor memory}
\State \textbf{return} $\widehat{a}^{(t)}_k,\, m$
\end{algorithmic}
\end{algorithm}

To implement this design, as shown in Algorithm~\ref{alg:abs_value_predictor}, at each training epoch~$t$, the predictor uses the reconstructed absolute gradient from the previous round, $\tilde{a}^{(t-1)}_k=|\tilde{g}^{(t-1)}_k|$. We first normalize $\tilde{a}^{(t-1)}_k$ using its mean and standard deviation to remove scale variation across epochs. The normalized value is then processed by an exponential moving average (EMA) as shown in line 3, which produces a smoothed prediction in the normalized space. This prediction is subsequently de-normalized using the mean and standard deviation of the current gradient $g_t$, yielding the predicted magnitude $\hat{a}^{(t)}_k$. Finally, the memory state is updated for the next step. For initialization, we set the memory $m=0$ when $t=1$.

This predictor is designed to be lightweight and online. The normalization and EMA update incur only $O(n)$ operations per epoch, where $n$ is the number of gradient elements, which is negligible compared to the cost of local training. Moreover, the predictor does not require storing a history of all previous gradients. Instead, it maintains only a single memory tensor $m$ of the same dimension as the gradient, making its memory overhead minimal. Normalization is critical in this setting. Without it, the dynamic range of gradients can drift significantly across epochs and layers, causing the EMA to lose accuracy. By normalizing before prediction and denormalizing after, the predictor adapts to the non-stationary nature of gradient statistics while retaining the smooth low-frequency trend. This combination of EMA and normalization enables accurate, stable, and efficient absolute-value prediction for compression. Importantly, by basing the prediction on reconstructed gradients rather than raw values, the predictor ensures that the client and server can remain fully synchronized using only the transmitted compressed payload (including $\sigma_{\text{curr}}$ and $\mu_{\text{curr}}$), without any need for additional side information. This combination of EMA and normalization enables accurate, stable, and efficient magnitude prediction for compression.

\subsection{Gradient Sign Predictor}
\label{subsection: sign prediction}

\begin{figure}[t]
    \centering
    \begin{subfigure}{0.48\linewidth}
        \centering
        \captionsetup{font=scriptsize}
        \includegraphics[width=\linewidth]{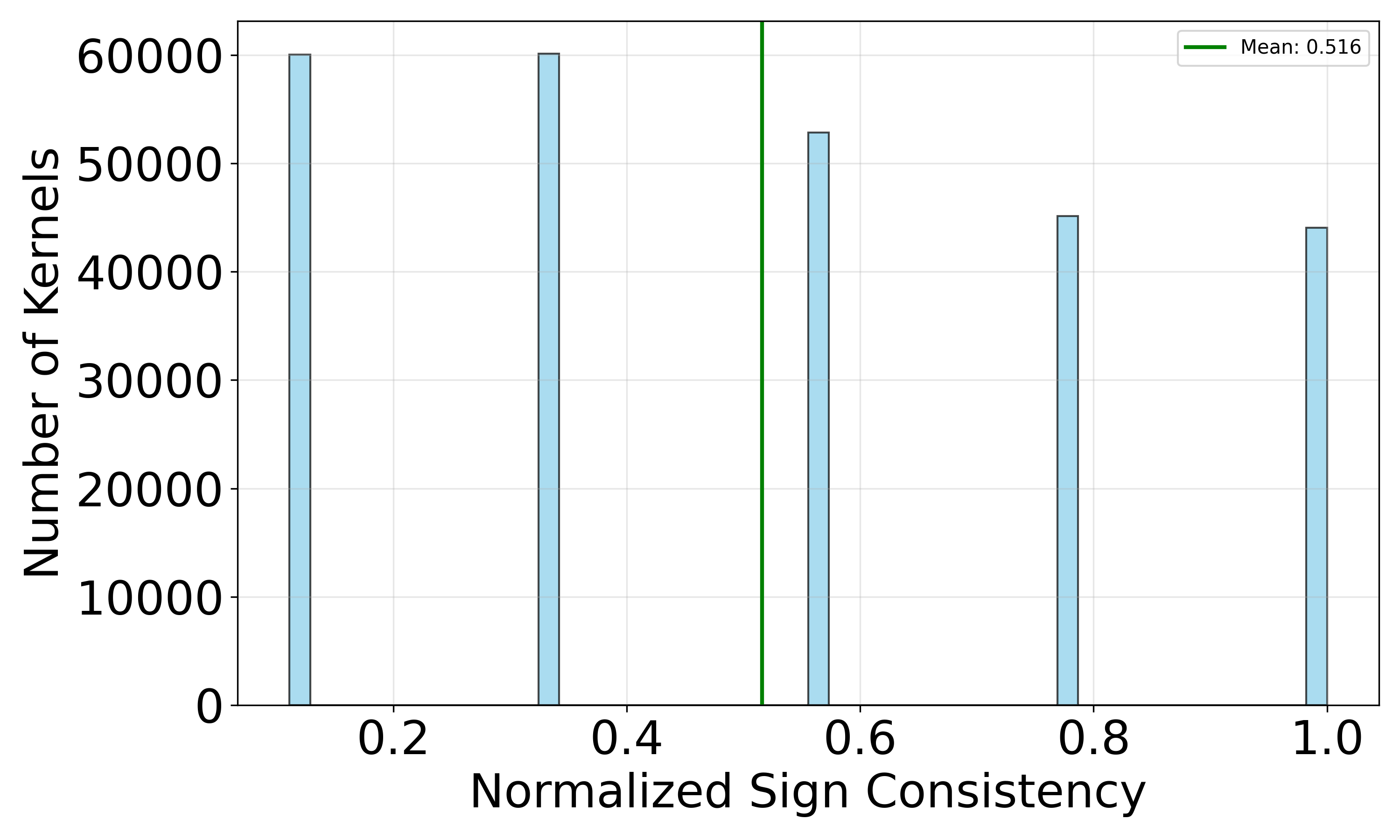}
        \caption{Conv layer sign consistency distribution}
        \label{fig:sub1}
    \end{subfigure}
    \hfill
    \begin{subfigure}{0.48\linewidth}
        \centering
        \captionsetup{font=scriptsize}
        \includegraphics[width=\linewidth]{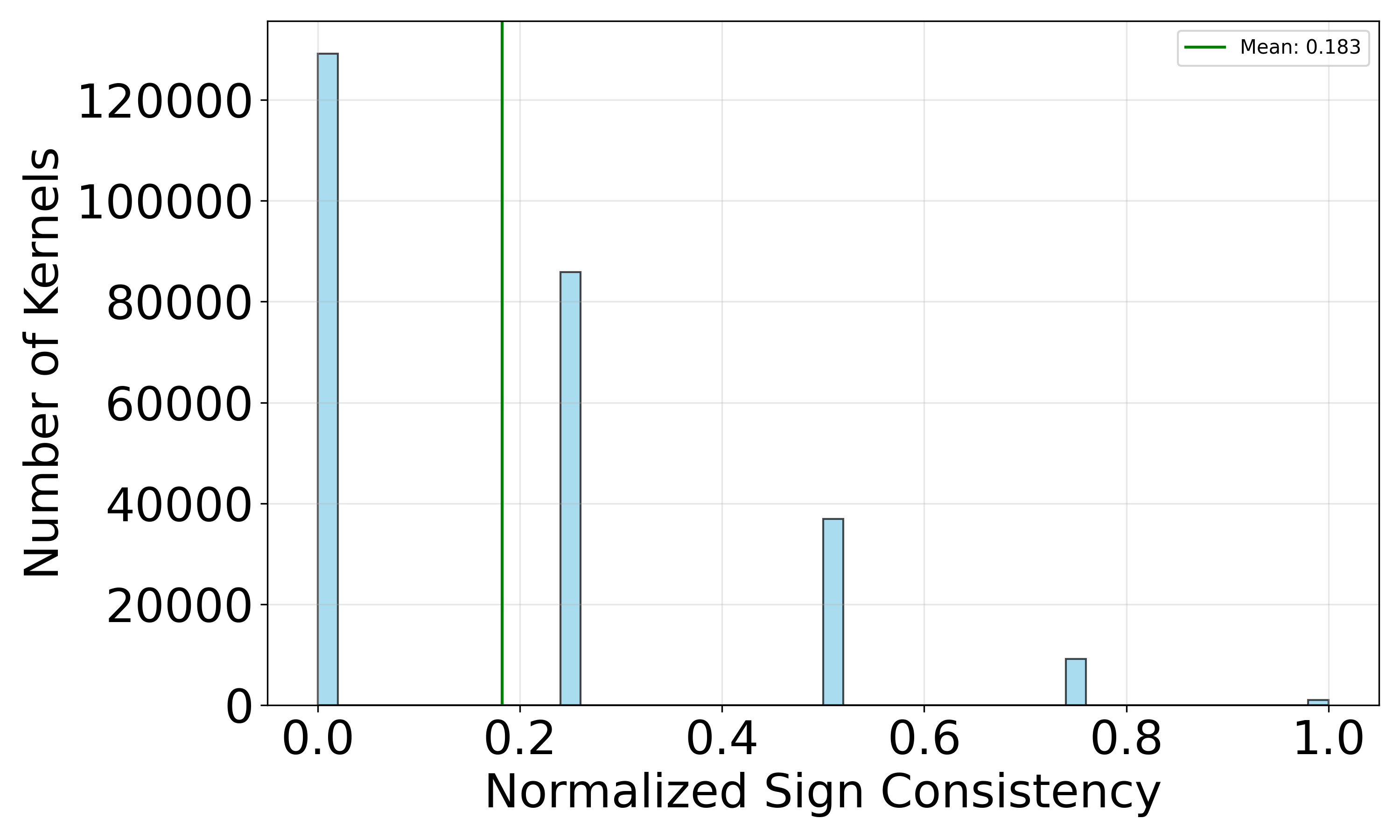}
        \caption{Random kernel sign consistency}
        \label{fig:sub2}
    \end{subfigure}

    \vspace{0.5em}  

    \begin{subfigure}{0.48\linewidth}
        \centering
        \captionsetup{font=scriptsize}
        \includegraphics[width=\linewidth]{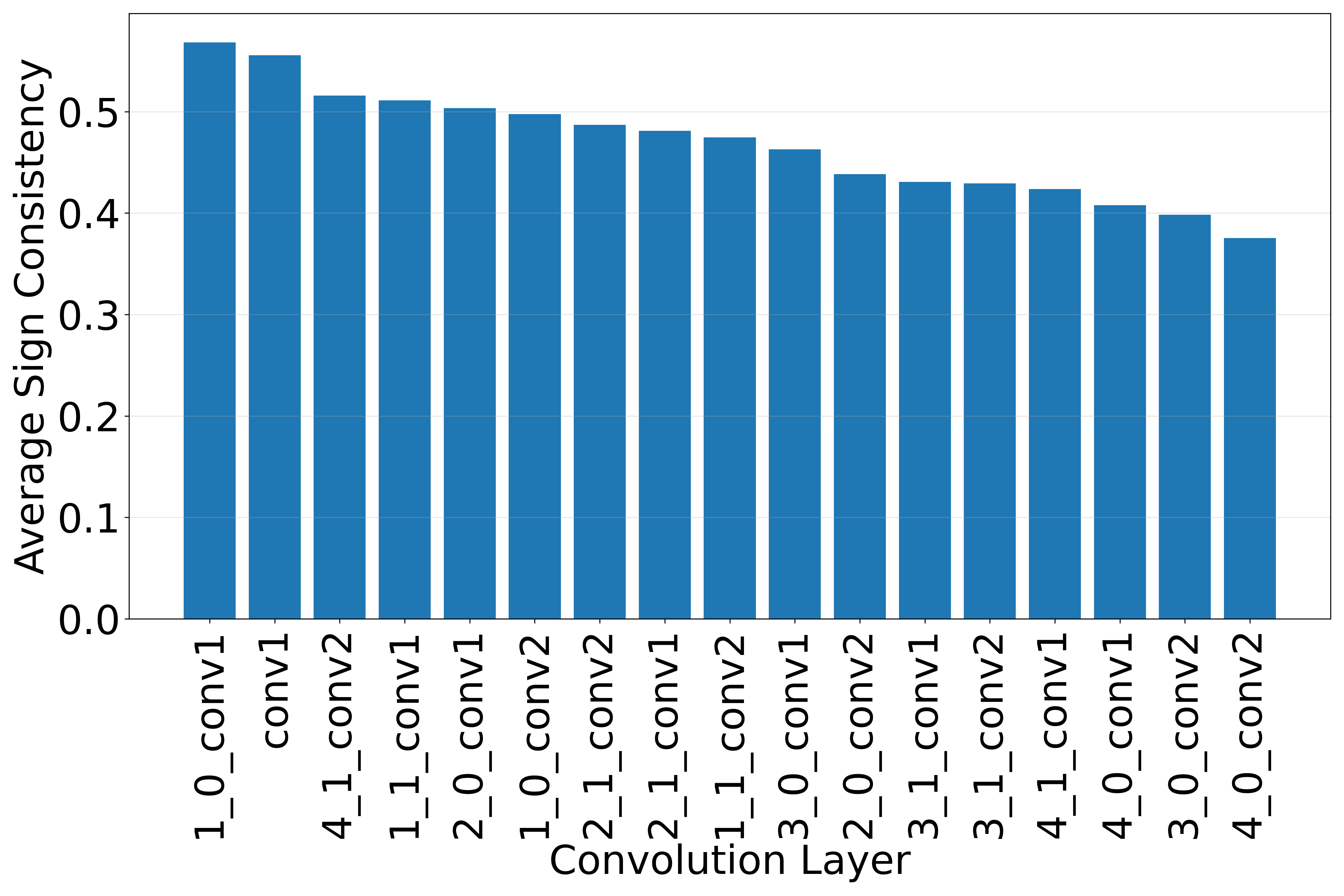}
        \caption{Average sign consistency at epoch 10}
        \label{fig:sub3}
    \end{subfigure}
    \hfill
    \begin{subfigure}{0.48\linewidth}
        \centering
        \captionsetup{font=scriptsize}
        \includegraphics[width=\linewidth]{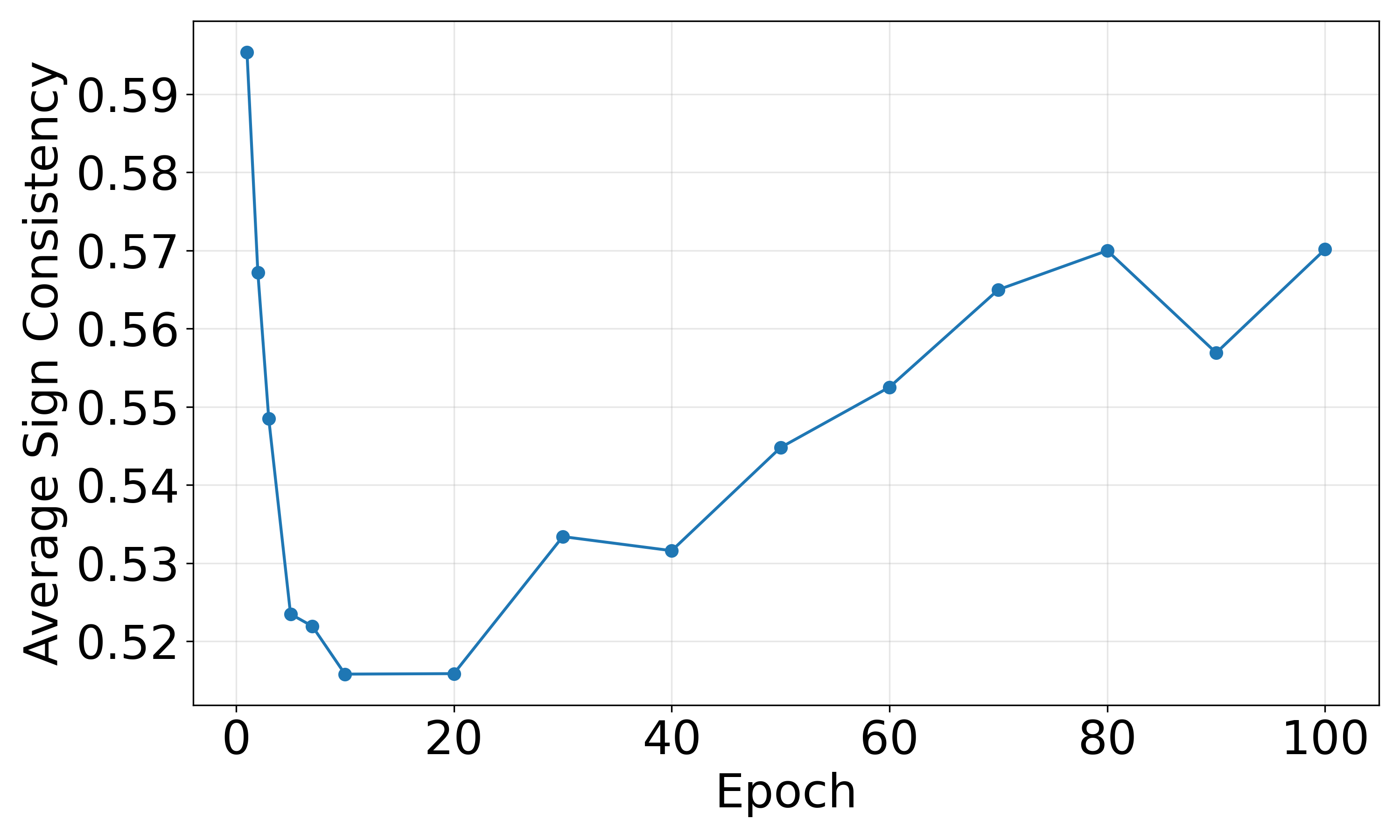}
        \caption{Sign consistency across training epoch}
        \label{fig:sub4}
    \end{subfigure}

    \caption{Kernel sign consistency statistics, including per-layer distribution and random baseline, average consistency across convolutional layers, and layer consistency across training epochs.}
    \label{fig:sign_consistency}
\end{figure}

In addition to the magnitude predictor, we also design a dedicated sign predictor. This predictor is built on two distinct cases depending on the training regime. In full-batch gradient descent, it is well-known that gradients tend to oscillate around the optimum with high temporal correlation. This oscillatory pattern implies that the sign of the current gradient can often be inferred from the previous round, potentially with a global flip depending on the gradient correlation. Thus, in this case, the predictor leverages the oscillation property directly to determine whether the sign should be preserved or flipped. For mini-batch training, however, such oscillation is much less stable, and the sign of individual gradients is highly noisy. Instead, we exploit a structural property of convolutional layers: within each convolutional kernel, most elements tend to share the same sign. To formalize this, we define the \emph{sign consistency} of a kernel $K$ of size $T$ as
\begin{equation}
    \text{Consistency}(K) = \frac{Max(P, N) + Z-\lceil T/2\rceil}{T-\lceil T/2\rceil} ,
    \label{eq:sign consistency}
\end{equation}
where $P$, $N$, and $Z$ denote the number of positive, negative, and zero elements in the kernel, respectively. This measure is normalized to the range $[0,1]$, with higher values indicating stronger agreement of element signs within the kernel. Intuitively, when a kernel’s consistency is close to one, nearly all elements exhibit the same sign while zeros are treated as neutral. In contrast, a value near zero suggests no dominant sign pattern.

To validate that this property is both meaningful and stable, we present the empirical evidence in Fig.~\ref{fig:sign_consistency}. Fig.~\ref{fig:sign_consistency} (a) shows the distribution of sign consistency for one convolutional layer of the model at a given epoch. For comparison, Fig.~\ref{fig:sign_consistency} (b) reports the same metric when kernels are randomly generated. The results clearly demonstrate that real kernels exhibit significantly higher sign consistency than random ones, indicating that gradients indeed possess a dominant sign structure within kernels. Fig.~\ref{fig:sign_consistency} (c) further shows the average sign consistency across different convolutional layers within the same epoch. The values are closely clustered, suggesting that this property is not confined to particular convolutional layers but is generally shared among them. Fig.~\ref{fig:sign_consistency} (d) plots the average sign consistency of a representative convolutional layer across training epochs. We observe that, although consistency can vary slightly during training (reaching its minimum at epoch 10 as we shown here), it remains consistently high overall. Together, these observations confirm that kernel-level sign consistency is a robust and stable feature of gradients throughout the training process.

\begin{algorithm}
\footnotesize
\caption{Gradient Sign Predictor}
\label{alg:sign}
\begin{algorithmic}[1]
\Statex \textbf{Input:} Current gradient $g^{(t)}_{k}$; reconstructed gradient from previous epoch $\tilde g^{(t-1)}_{k}$; consistency threshold $\tau \in [0,1]$; optional full-batch-GD flag.
\Statex \textbf{Output:} Elementwise predicted sign tensor $S^{(t)}_{k}$.
\State Initialize $S^{(t)}_{k}$ as a zero tensor with the same shape as $g^{(t)}_{k}$.
\If{full-batch GD flag is true}
   \State $c \gets \textsc{GradientCorrelation}(\tilde g^{(t-1)}_{k}, g^{(t)}_{k})$
   \State $S^{(t)}_{k} = Sign(c) \times S^{(t-1)}_{k}$ \Comment{predict base on the last gradient}
\Else
    \For{each kernel $K^{(t)}_{o,i}$ in $g^{(t)}_{k}$}
       \State $\text{consistency} \gets \textsc{SignConsistency}(K^{(t)}_{o,i})$
       \If{$\text{consistency} \ge \tau$} \Comment{check whether the kernel is predictable}
          \State $\text{dominant sign} \gets \textsc{GetDominantSign}(K^{(t)}_{o,i})$
          \State $S^{(t)}_{k}[o,i,:,:] \gets \text{dominant sign}$ \Comment{predict with dominant sign}
       \Else
          \State $S^{(t)}_{k}[o,i,:,:] \gets 0$ \Comment{no prediction for this kernel}
       \EndIf
    \EndFor
\EndIf

\State \textbf{return} $S^{(t)}_{k}$
\end{algorithmic}
\end{algorithm}
As shown in Algorithm~\ref{alg:sign}, the gradient sign predictor takes as input the current gradient tensor $g^{(t)}_k$, the previous epoch's reconstructed gradient $\tilde g^{(t-1)}_k$, a kernel-level consistency threshold $\tau$, and an optional flag indicating full-batch training. The output is an elementwise sign tensor $S^{(t)}_k$ used by the downstream value predictor to form the gradient prediction $\widehat g^{(t)}_k = S^{(t)}_k \odot \widehat a^{(t)}_k.$ Under full-batch training, as shown in line 2-4, a global oscillation check is applied by computing a scalar gradient correlation $c$. If $c < 0$, the predicted signs in $S^{(t)}_k$ are directly assigned as the negation of $S^{(t-1)}_k$, reflecting the expected reversal of gradient direction. Otherwise, $S^{(t)}_k$ inherits the previous signs without modification. For mini-batch training, as shown in line 6-14, the algorithm scans each convolutional kernel $K^{(t)}_{o,i}\subset g^{(t)}_k$ where $o$ and $i$ are the index of output and input channel, computes its sign consistency, and, if the score exceeds $\tau$, assigns the kernel's dominant sign to all its elements; otherwise, no sign is predicted for that kernel and zeros are left in $S^{(t)}_k$ to indicate ``no-op''. 

This sign predictor is lightweight and requires only simple counting and thresholding operations without introducing additional heavy computations. The overall complexity is $O(n)$ for $n$ gradient elements, which is negligible compared to the cost of local training. In the full-batch case the oscillation correction relies solely on the correlation between two consecutive gradients so only one bit to indicate flip or not and no extra communication between client and server is needed. In the mini-batch case kernel-level prediction does require transmitting bitmaps to indicate which kernels are predicted and the sign of it, but the size of this bitmap is tiny relative to the gradient data and its overhead is negligible. Next, we describe how the bitmap is generated and efficiently handled within our compression pipeline.

\subsection{Auxiliary Metadata}
In gradient sign prediction, unlike in the full-batch case where signs can be inferred purely from temporal correlation of historical gradients and only need 1 bit to indicate the flipping, the kernel consistency based predictor requires transmitting a small amount of side information to ensure consistent and efficient prediction on the server side. To this end, we introduce a bitmap mechanism that explicitly records which kernels are predicted and what their dominant signs are.

\begin{figure}
    \centering
    \includegraphics[width=\linewidth]{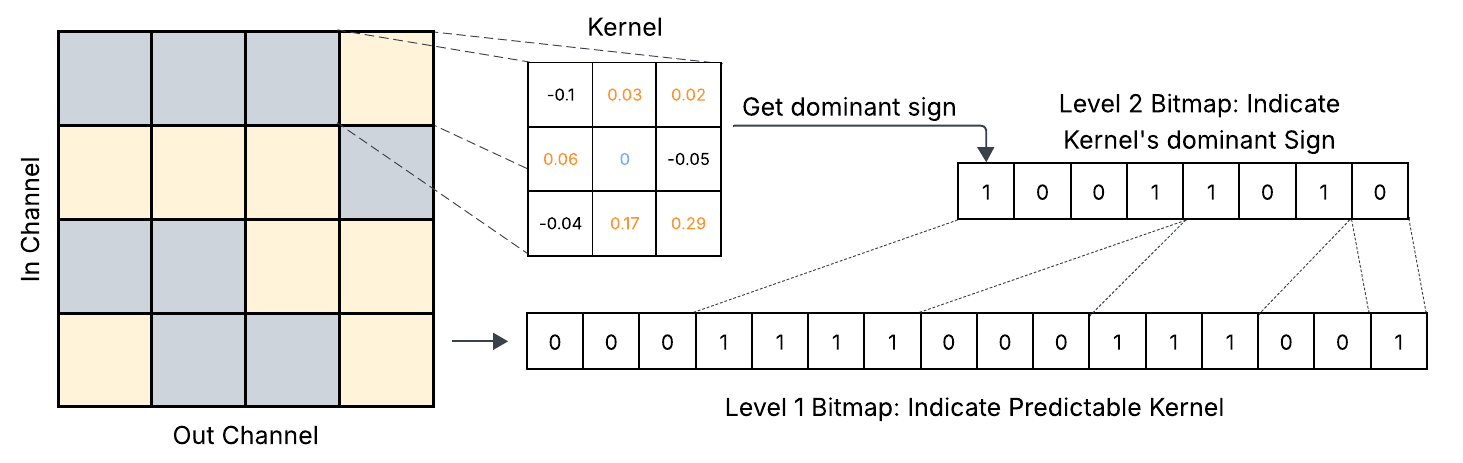}
    \caption{The illustration of the two-level bitmap structure. }
    \label{fig:bitmap}
\end{figure}

We design a two-level bitmap structure, illustrated in Figure ~\ref{fig:bitmap}. The first-level bitmap indicates which kernels are selected for sign prediction. Given a sign consistency threshold and kernel size, we compute the corresponding threshold on the number of dominant-sign elements within a kernel. If the number of dominant-sign elements exceeds this threshold, the kernel is predicted using its dominant sign, and the corresponding position in the first-level bitmap is set to one, otherwise zero. The second-level bitmap is only generated for those kernels whose first-level entry is 1. It records the dominant sign of each predictable kernel, where 1 represents positive and 0 represents negative. 

The generated bitmaps are passed through entropy coding and then bundled together with the quantized residuals into the lossless compressor (e.g., Zstd or Blosc). Although this introduces an additional component to the payload, the cost is negligible. The relative size of the bitmap compared to the original data is $\frac{1+P}{b \cdot K \cdot R},$ where $b$ is the number of bits used for the original gradient data type, $K$ is the kernel size, and $R$ is the lossless compression ratio, and $P$ is the prediction ratio, which denotes the fraction of kernels whose sign consistency exceeds the chosen threshold. For example, when the gradient is stored in 32-bit floating-point format, $K=3\times 3$, $R=1.2$, and $p=0.6$, the bitmap is only $0.46\%$ of the original size. This tiny overhead confirms that incorporating the bitmap does not compromise the overall compression ratio.

\begin{algorithm}
\footnotesize
\caption{CompressOneRound$(\{g^{(t)}_{k,\ell}\}_{\ell=1}^{L})$ for client $k$}
\label{alg:compress_epoch_revised}
\begin{algorithmic}[1]
\Statex \textbf{Input} layer-$l$ epoch-$t$ gradients $\{g^{(t)}_{k,\ell}\}$; previous reconstructed gradients $\{\tilde g^{(t-1)}_{k,\ell}\}$; parameters $(\alpha,\tau,\textsc{FullBatchGD},\textsc{ErrMode},\Delta,T_{\textsc{lossy}})$
\Statex \textbf{Output} compressed payload $\mathcal{C}^{(t)}_k$
\State $\mathcal{C}^{(t)}_k \gets \emptyset$
\For{$\ell=1$ to $L$}
  \If{$\text{numel}(g^{(t)}_{k,\ell}) \le T_{\textsc{lossy}}$} \Comment{lossless for small layers}
     \State $\mathcal{B}_\ell \gets \textsc{LosslessCompress}(g^{(t)}_{k,\ell})$; append to $\mathcal{C}^{(t)}_k$; 
     \State \textbf{continue}
  \EndIf
  \State $a^{(t)}_{k,\ell} \gets |g^{(t)}_{k,\ell}|$; \quad $\tilde a^{(t-1)}_{k,\ell} \gets |\tilde g^{(t-1)}_{k,\ell}|$
  \State$\mu_{\text{curr}},\sigma_{\text{curr}} \gets \big(\operatorname{mean}(a^{(t)}_{k,\ell}),\,\operatorname{std}(a^{(t)}_{k,\ell})\big)$
  \State $\widehat a^{(t)}_{k,\ell},\, m_\ell \gets \textsc{MagnitudePredict}(\tilde a^{(t-1)}_{k,\ell};\, \mu_{\text{curr}},\sigma_{\text{curr}},\alpha)$
    \State $S^{(t)}_{k,\ell}, b_{\ell} \gets \textsc{SignPredict}(g^{(t)}_{k,\ell};\tilde{g}^{(t-1)}_{k,\ell} \tau, \textsc{FullBatchGD})$

  \State $\widehat g^{(t)}_{k,\ell} \gets S^{(t)}_{k,\ell} \odot \widehat a^{(t)}_{k,\ell}$ \Comment{form prediction}
  \State $e^{(t)}_{k,\ell} \gets g^{(t)}_{k,\ell} - \widehat g^{(t)}_{k,\ell}$ \Comment{get residual}
  \State $q^{(t)}_{k,\ell} \gets \textsc{EBQuantize}(e^{(t)}_{k,\ell}; \textsc{ErrMode}, \Delta)$
  \State $h^{(t)}_{k,\ell} \gets \textsc{EntropyEncode}(q^{(t)}_{k,\ell})$
  \State $\mathcal{B}_\ell \gets \textsc{LosslessCompress}(\mu_{\text{curr}},\sigma_{\text{curr}};\; h^{(t)}_{k,\ell}, b_\ell)$

  \State append $\mathcal{B}_\ell$ to $\mathcal{C}^{(t)}_k$
\EndFor
\State \textbf{return} $\mathcal{C}^{(t)}_k$
\end{algorithmic}
\end{algorithm}

\begin{algorithm}[t]
\footnotesize
\caption{DecompressOneRound$(\mathcal{C}^{(t)}_k)$ at server}
\label{alg:decompress_epoch_match}
\begin{algorithmic}[1]
\Statex \textbf{Input} payload $\mathcal{C}^{(t)}_k$; previous reconstructed $\{\tilde g^{(t-1)}_{k,\ell}\}$;
parameters $(\alpha,\tau,\textsc{ErrMode},\Delta,T_{\textsc{lossy}})$
\Statex \textbf{Output} reconstructed gradients $\{\tilde g^{(t)}_{k,\ell}\}$
\For{each layer blob $\mathcal{B}_\ell$ in $\mathcal{C}^{(t)}_k$}
  \State $(\text{tag}, \cdot) \gets \textsc{PeekHeader}(\mathcal{B}_\ell)$ \Comment{tag indicates lossy or lossless}
  \If{$\text{tag}=\textsc{LosslessOnly}$}
     \State $\tilde g^{(t)}_{k,\ell} \gets \textsc{LosslessDecompress}(\mathcal{B}_\ell)$
     \State \textbf{continue}
  \EndIf
  \State  $(\mu_{\text{curr}},\sigma_{\text{curr}}), h^{(t)}_{k,\ell}, b_\ell \gets \textsc{LosslessDecompress}(\mathcal{B}_\ell)$
  \State $q^{(t)}_{k,\ell} \gets \textsc{EntropyDecode}(h^{(t)}_{k,\ell})$
  \State $\tilde a^{(t-1)}_{k,\ell} \gets |\tilde g^{(t-1)}_{k,\ell}|$
  \State $\widehat a^{(t)}_{k,\ell},\, m_\ell \gets \textsc{MagnitudePredict}(\tilde a^{(t-1)}_{k,\ell};\, \mu_{\text{curr}},\sigma_{\text{curr}},\alpha)$
  \State $S^{(t)}_{k,\ell} \gets \textsc{ReconstructSignFromBitmap}(b_\ell,\ \tilde g^{(t-1)}_{k,\ell})$
  \State $\widehat g^{(t)}_{k,\ell} \gets S^{(t)}_{k,\ell} \odot \widehat a^{(t)}_{k,\ell}$ \Comment{form prediction}
  \State $e^{(t)}_{k,\ell} \gets \textsc{EBDequantize}(q^{(t)}_{k,\ell};\ \textsc{ErrMode},\ \Delta)$
  \State $\tilde g^{(t)}_{k,\ell} \gets \widehat g^{(t)}_{k,\ell} + e^{(t)}_{k,\ell}$ \Comment{reconstruct gradient}
\EndFor
\State \textbf{return} $\{\tilde g^{(t)}_{k,\ell}\}$
\end{algorithmic}
\end{algorithm}

\subsection{Detailed Implementation}
The complete workflow of our compressor is summarized in Algorithm~\ref{alg:compress_epoch_revised} and Algorithm~\ref{alg:decompress_epoch_match}. In Algorithm~\ref{alg:compress_epoch_revised}, line 2-5 illustrate that on the client side, each layer is first checked against the lossy–lossless threshold. Small layers are stored directly with lossless compression, while larger layers undergo prediction-based lossy compression. For these layers, we apply the magnitude predictor using the previous reconstructed gradients and the current mean and standard deviation as shown in line 7-9, and the sign predictor using either oscillation correlation or kernel-level consistency in line 10. The predicted gradients are first used to compute the residuals against the true gradients. These residuals are then quantized, entropy-encoded, and finally packaged together with the bitmap through a lossless compressor as shown in line 11-16.

In Algorithm~\ref{alg:decompress_epoch_match}, the decompression process on the server side mirrors the client workflow to ensure deterministic reconstruction. Each layer blob is first decoded to check whether lossless or lossy compression was applied as shown in lines 2-4. For lossy-compressed layers, the transmitted mean and standard deviation are decompressed and reused to reproduce the magnitude prediction as shown in lines 7-10. The sign information is reconstructed either from a single global flip bit in the full-batch setting or from a compact kernel-level bitmap in the mini-batch case as shown in line 11. The predicted magnitude and sign are then combined to regenerate the gradient estimate, and the dequantized residual is added back to reconstruct the original gradient with bounded error as shown in lines 12-14. This process ensures full consistency between client and server while maintaining high accuracy and minimal side-information overhead.

\section{Evaluation}
In this section we will provide details about evaluation method and analyze the results.

\subsection{Platform Integration and Experiment Setup}
We conduct our experiments on APPFL (v1.2.1)~\cite{ryu2022appfl}, an open-source Python framework specifically designed for research in federated and privacy-preserving learning. The framework provides modular components for learning algorithms, compressors, communication protocols, and data pipelines, which makes it well-suited for integrating custom compression modules. APPFL supports PyTorch models directly, while distributed training is implemented using gRPC for client–server communication and MPI through \texttt{mpi4py}~\cite{dalcin2021mpi4py} for scalable parallel execution. This design enables us to simulate realistic FL workloads under controlled experimental settings.

For the federated aggregation algorithm, we adopt FedAvg~\cite{mcmahan2017communication}, the canonical method for FL training. In FedAvg, each client performs local gradient descent on its private data and periodically transmits gradients to the server, which then aggregates them via weighted averaging to update the global model. We choose FedAvg because of its simplicity, strong empirical performance, and wide adoption, which provides a solid and representative baseline for evaluating the impact of our compression method.

In our experiments, our proposed EBLC is configured with the same quantizer, entropy coder, and lossless backend as SZ3, because this configuration is generally optimal in most cases. This setup ensures a fair comparison with the SZ3 baseline and allows the effectiveness of our design to be clearly observed.

All experiments are conducted on the Polaris supercomputer at Argonne National Laboratory. Each compute node on Polaris is equipped with an AMD EPYC Milan CPU (32 cores), 512 GB of system memory, and four NVIDIA A100 GPUs (40 GB HBM2e each), connected with NVLink. Nodes are interconnected via a high-speed HPE Slingshot network. This large-scale GPU-accelerated environment allows us to emulate federated training with realistic communication costs at scale.

\subsection{Models and Datasets}
To comprehensively evaluate the effectiveness of our proposed compressor, we select a diverse set of convolutional neural networks (CNNs) and datasets that differ significantly in size, depth, and structural complexity. Our design goal is to reduce communication overhead in FL
without compromising model accuracy. Thus, it is essential to verify that the proposed method is not tied to a specific model scale or dataset property. 

\begin{table}[h]
\vspace{-6pt}
\centering
\caption{CNN models used in evaluation}
\vspace{-6pt}
\label{tab:models}
\resizebox{0.7\linewidth}{!}{
\begin{tabular}{lcccc}
\toprule
\textbf{Model} & \textbf{Layers} & \textbf{Params (M)} & \textbf{Size (MB)} \\
\midrule
ResNet-18~\cite{he2016deep}     & 18 & 11.7  & 44.6 \\
ResNet-34~\cite{he2016deep}     & 34 & 21.8  & 87.2 \\
Inception V1~\cite{szegedy2015going}  & 22 & 6.6   & 26.4 \\
Inception V3~\cite{szegedy2016rethinking}  & 48 & 23.9  & 95.6 \\
\bottomrule
\end{tabular}
}
\vspace{-6pt}
\end{table}

For model selection, we aim to cover variations in both scale and architecture. ResNet-18 and ResNet-34 belong to the same residual network family but differ in depth and parameter size, allowing us to study performance across different model scales. 
Inception V1 and V3, by contrast, employ multi-branch convolutional structures instead of residual connections, while maintaining comparable parameter sizes to the ResNets. This combination enables us to compare architectures of similar scale but different design principles. 
Together, these choices ensure that our evaluation demonstrates the generalization ability of the proposed compressor across both model size and architectural design. 

\begin{table}[h]
\vspace{-6pt}
\centering
\caption{Datasets used in evaluation}
\vspace{-6pt}
\label{tab:datasets}
\resizebox{0.8\linewidth}{!}{
\begin{tabular}{lcccc}
\toprule
\textbf{Dataset} & \textbf{\#Images} & \textbf{Resolution} & \textbf{Classes} & \textbf{Channels} \\
\midrule
Fashion-MNIST~\cite{xiao2017fashion}  & 70,000  & $28 \times 28$   & 10  & 1 \\
CIFAR-10~\cite{krizhevsky2009learning}       & 60,000  & $32 \times 32$   & 10  & 3 \\
Caltech101~\cite{fei2004learning}    & 9,000 & $224 \times 224$ & 101 & 3 \\
\bottomrule
\end{tabular}
}
\end{table}

\begin{table*}[t]
\centering
\caption{Comparison of compression ratios  between Ours, SZ3 and QSGD across models and datasets}
\vspace{-1em}
\label{tab:compression_ratios}
\renewcommand{\arraystretch}{0.8}
\resizebox{0.8\linewidth}{!}{
\begin{tabular}{cc cccc cccc cccc}
\toprule
Dataset
 & \multicolumn{5}{c}{CIFAR-10} 
 & \multicolumn{4}{c}{Caltech101} 
 & \multicolumn{4}{c}{Fashion-MNIST} \\
\midrule
REL Error Bound & 
 & $10^{-3}$ & $10^{-2}$ & $3\times10^{-2}$ & $5\times10^{-2}$
 & $10^{-3}$ & $10^{-2}$ & $3\times10^{-2}$ & $5\times10^{-2}$
 & $10^{-3}$ & $10^{-2}$ & $3\times10^{-2}$ & $5\times10^{-2}$ \\
\midrule
\multirow{2}{*}{ResNet-18} 
 & Ours & \textbf{5.43} & \textbf{9.78} & \textbf{16.80} & \textbf{28.97} & \textbf{4.79} & \textbf{9.82} & \textbf{21.94} & \textbf{28.66} & \textbf{4.95} & \textbf{9.55} & \textbf{16.96} & \textbf{28.04} \\
 & SZ3  & 4.92 & 8.52 & 13.68 & 23.80 & 4.46 & 9.00 & 18.79 & 26.69 & 4.48 & 8.20 & 13.63 & 20.64 \\
 & QSGD & 5.01 & 8.62 & 12.27 & 13.74 & 4.55 & 7.97 & 11.98 & 13.59 & 4.63 & 7.56 & 11.35 & 13.21 \\
\midrule
\multirow{2}{*}{ResNet-34} 
 & Ours & \textbf{5.34} & \textbf{11.95} & \textbf{24.09} & \textbf{28.46} & \textbf{6.33} & \textbf{18.14} & \textbf{37.56} & \textbf{54.06} & \textbf{5.05} & \textbf{11.50} & \textbf{21.48} & \textbf{26.77} \\
 & SZ3  & 4.91 & 10.12 & 18.31 & 21.57 & 5.98 & 16.10 & 34.84 & 50.38 & 4.54 & 9.25 & 15.50 & 17.53 \\
 & QSGD & 5.12 & 8.04 & 11.95 & 13.23 & 6.02 & 10.27 & 13.70 & 14.50 & 4.77 & 8.19 & 11.95 & 14.32 \\
\midrule
\multirow{2}{*}{Inception-V1} 
 & Ours & \textbf{4.27} & \textbf{8.19} & \textbf{14.21} & \textbf{16.75} & \textbf{4.15} & \textbf{7.71} & \textbf{11.06} & \textbf{14.71} & \textbf{4.20} & \textbf{7.83} & \textbf{13.50} & \textbf{17.67} \\
 & SZ3  & 4.03 & 7.29 & 12.03 & 15.31 & 3.82 & 6.60 & 9.66 & 12.76 & 4.03 & 7.54 & 12.84 & 16.38 \\
 & QSGD & 3.89 & 6.25 & 9.64 & 11.64 & 3.94 & 6.21 & 9.24 & 11.34 & 3.86 & 5.93 & 8.96 & 12.28 \\
\midrule
\multirow{2}{*}{Inception-V3} 
 & Ours & \textbf{4.73} & \textbf{9.31} & \textbf{14.98} & \textbf{18.64} & \textbf{5.68} & \textbf{11.81} & \textbf{23.40} & \textbf{32.10} & \textbf{4.78} & \textbf{8.39} & \textbf{12.62} & \textbf{27.20} \\
 & SZ3  & 4.15 & 7.47 & 11.49 & 17.39 & 5.53 & 10.77 & 21.53 & 30.54 & 4.32 & 7.07 & 9.87 & 19.55 \\
 & QSGD & 4.30 & 7.78 & 11.61 & 13.21 & 4.89 & 8.90 & 12.15 & 13.40 & 4.45 & 7.69 & 11.17 & 13.73 \\
\bottomrule
\end{tabular}
}
\end{table*}

\begin{figure*}
    \centering
    \includegraphics[width=0.8\linewidth]{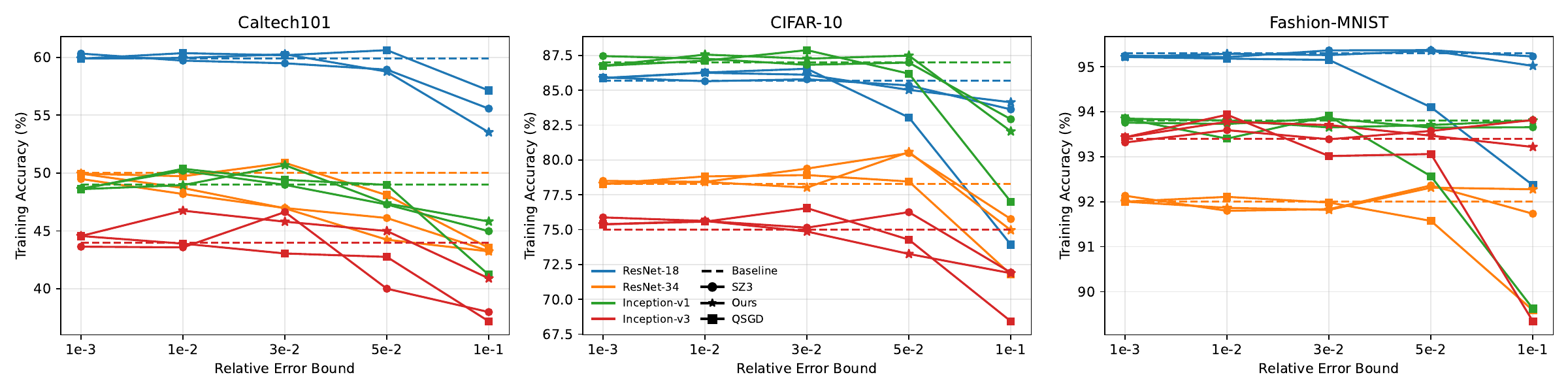}
    \vspace{-1.5em}
    \caption{Comparison of training accuracy  between Ours, SZ3 and QSGD across models and datasets}
    \label{fig:training acc}
\end{figure*}

For datasets, we adopt three widely used datasets with distinct characteristics. Fashion-MNIST is a simple grayscale dataset with low resolution , 
providing a controlled setting to test compression under minimal data complexity. CIFAR-10 increases the difficulty with RGB images of higher resolution, 
commonly used for evaluating vision models in FL. Finally, Caltech101 is a much more challenging dataset with 101 categories and variable resolution images. This dataset better reflects real-world scenarios where both resolution and class diversity are higher. 

Together, these models and datasets allow us to rigorously assess whether our method consistently achieves higher compression ratios while preserving accuracy, across different network architectures and data regimes.

\subsection{Model-Wise Performance Analyses}
We evaluate the proposed compressor across different models and datasets under a consistent experimental protocol. For each model-dataset combination, we train for 10 epochs and record both the average model-wise compression ratio and the final training accuracy. The compressors operate under relative error bounds ranging from $10^{-3}$ to $10^{-1}$. And we compare our proposed method with the state-of-the-art SZ3 compressor to directly measure the improvement of out design. In addition to SZ3, we also include QSGD as a widely adopted quantization-based baseline. Since QSGD does not natively provide error-bound control, we configure it using a fixed number of quantization levels to approximate the error bounds used in our experiments. Specifically, we set the quantization bit-width to \{10, 7, 5, 4, 3\}, which correspond to bin counts comparable to those implied by relative error bounds of $10^{-3}$, $10^{-2}$, $3\times10^{-2}$, $5\times10^{-2}$, and $10^{-1}$, respectively. This setup enables a fair side-by-side comparison between QSGD and our error-bounded lossy compressor under similar compression ratios.

Table~\ref{tab:compression_ratios} presents the achieved compression ratios with different combinations of models and datasets under various relative error bounds. From the table, we observe that our method consistently achieves higher compression ratios than SZ3 and QSGD. For example, on ResNet-34 with CIFAR-10 at a relative error bound of $3\times10^{-2}$, QSGD achieves a compression ratio of 11.95$\times$, SZ3 attains a compression ratio of 18.31$\times$, while our method reaches 24.09$\times$, which corresponds to a 31.56\% improvement compared to SZ3. Similar improvements can be observed in other model–dataset combinations, and overall, our method surpasses SZ3 and QSGD across all settings, with the maximum improvement over SZ3 reaching 52.67\%.

Another important observation is the impact of the error bound on compression gains. As the error bound increases from $10^{-3}$ to $3\times10^{-2}$, the relative advantage of our method over SZ3 expands steadily. This is because our predictors are not perfectly accurate, but they significantly reduce the variance of residuals, causing them to concentrate more tightly around zero. In this regime, the quantized residuals exhibit lower entropy, and entropy coding becomes more effective, leading to higher compression ratios. However, beyond this range, the improvements become less consistent. One reason is that as the error bound grows larger, the quantization bin size may already cover most of the concentrated residuals, leaving little room for further entropy reduction. Another reason is that the bitmap overhead, although small in absolute terms, remains fixed in size. As overall compression ratios increase, this fixed cost becomes more pronounced relative to the payload, partially offsetting the gains. Consequently, the improvement ratio may plateau or even slightly decline at higher error bounds.

In addition to the effect of the error bound, we also examine how model architecture and dataset characteristics influence compression performance. The overall model size has only a minor impact on the improvement ratio, but the gain varies slightly across different architectures. Residual-based networks such as ResNet tend to benefit more from our method than multi-branch architectures like Inception, as the gradient distributions in residual networks exhibit stronger temporal pattern that aligns well with our predictor design. In contrast, the choice of dataset has a more pronounced effect. Simpler datasets, such as Fashion-MNIST, yield more stable and predictable gradient patterns, resulting in larger compression gains. As the dataset complexity increases, for example in Caltech101, the gradient distributions become less regular and harder to predict, slightly reducing the relative advantage. Nevertheless, even on the most complex datasets, our compressor consistently provides a significant improvement over SZ3, demonstrating its robustness and adaptability across diverse training conditions.

Figure~\ref{fig:training acc} illustrates the training accuracy under different relative error bounds across various model–dataset combinations. The dashed lines represent the uncompressed baselines, while the star, dot and square markers correspond to our method, SZ3 and QSGD, respectively. Combined with Table \ref{tab:compression_ratios}, we observe that under the same error bound, our method consistently achieves higher compression ratios than SZ3 and QSGD, while maintaining similar accuracy to SZ3 and clearly outperforming QSGD. Moreover, As the error bound increases, compression ratios improve accordingly, but excessive bounds lead to noticeable accuracy degradation.

To further analyze this behavior, Figure~\ref{fig:training acc} reveals that the sensitivity of training accuracy to compression noise varies mainly across datasets rather than model architectures. Simpler datasets such as Fashion-MNIST are more tolerant: both our method and SZ3 retain nearly unchanged accuracy when the error bound is below $5\times10^{-2}$, whereas QSGD starts to degrade below 4 bits. A similar trend appears on CIFAR-10, where accuracy drops remain within 0.5\% up to the same threshold. More complex datasets like Caltech101 are less robust—acceptable accuracy is preserved only when the error bound is below $3\times10^{-2}$ for our method and SZ3, and below 5 bits for QSGD. Even under these tighter constraints, our method still yields higher compression ratios, demonstrating its strong efficiency–fidelity trade-off.

Overall, within the error bound range of $3\times10^{-2}$ to $5\times10^{-2}$, which typically represents the balance point between model utility and compression efficiency, our method achieves up to a 52.67\% improvement in compression ratio over SZ3. Since SZ3 is generally regarded as near-optimal, this substantial gain demonstrates the strong effectiveness of our design.
\begin{figure}
    \centering
    \includegraphics[width=\linewidth]{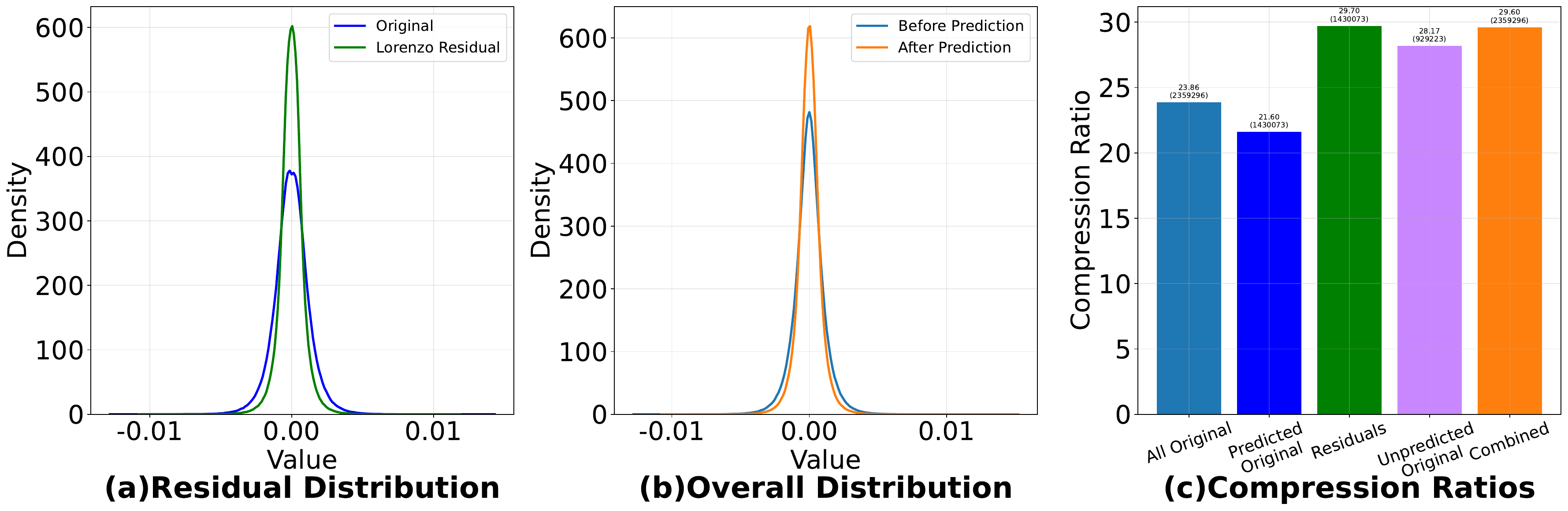}
    \vspace{-2em}
    \caption{Convolutional layer compression analysis. (a) Comparison of the distribution between predicted residuals and corresponding original parts. (b) Comparison of the distribution between overall compression data and original data. (c) Compression ratio for each part.}
    \vspace{-1.5em}
    \label{fig:conv layer comp analysis}
\end{figure}

\begin{table*}[t]
\centering
\caption{Comparison of compression ratios and prediction statistics across different kernel sizes}
\label{tab:kernel_analysis}
\resizebox{0.8\linewidth}{!}{
\begin{tabular}{ccccccccc}
\toprule
\multirow{2}{*}{Kernel Size} &
\multicolumn{5}{c}{Compression Ratio (CR)} &
\multicolumn{3}{c}{Prediction Statistics} \\
\cmidrule(lr){2-6} \cmidrule(lr){7-9}
 & All (SZ3) & Pred. (SZ3) & Residual (Ours) & Unpredicted & Combined (Ours)
 & Predict. Ratio & Sign Mismatch & Bitmap Overhead \\
\midrule
$3\times3$ & 23.86 & 21.60 & 29.70 & 28.17 & 29.60 & 60.60\% & 9.70\% & 12.07\% \\
$5\times5$ & 40.79 & 35.85 & 54.88 & 51.97 & 52.99 & 56.40\% & 10.33\% & 7.95\% \\
$7\times7$ & 46.64 & 42.11 & 57.90 & 48.08 & 51.36 & 30.40\% & 15.48\% & 3.43\% \\
\bottomrule
\end{tabular}
}
\end{table*}

\begin{figure*}[t]
    \centering
    \begin{subfigure}[b]{0.95\linewidth}
        \centering
        \includegraphics[width=\linewidth]{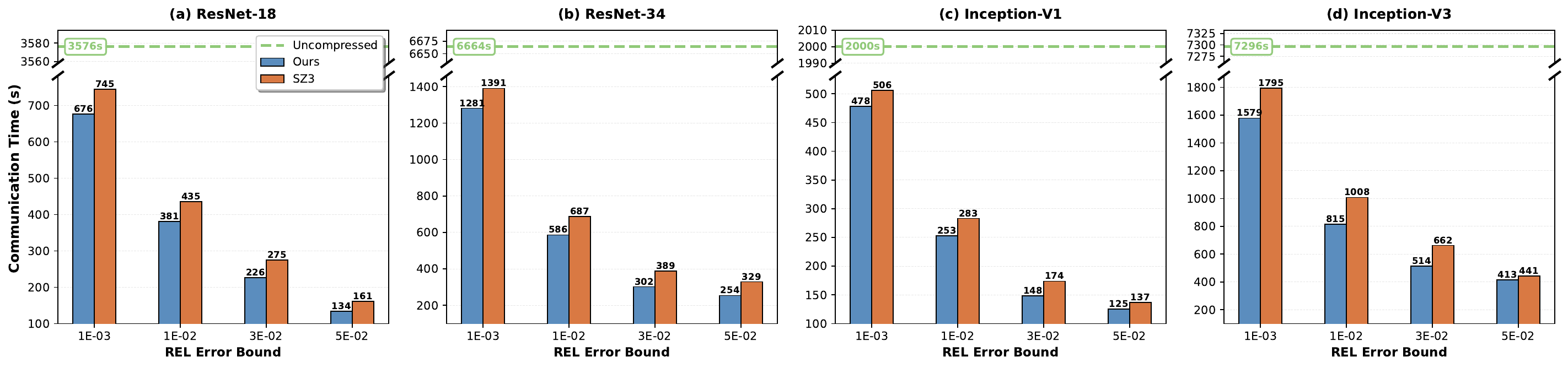}
        \label{fig:comm_time_over_eb}
    \end{subfigure}

    \vspace{-1.5em}

    \begin{subfigure}[b]{0.95\linewidth}
        \centering
        \includegraphics[width=\linewidth]{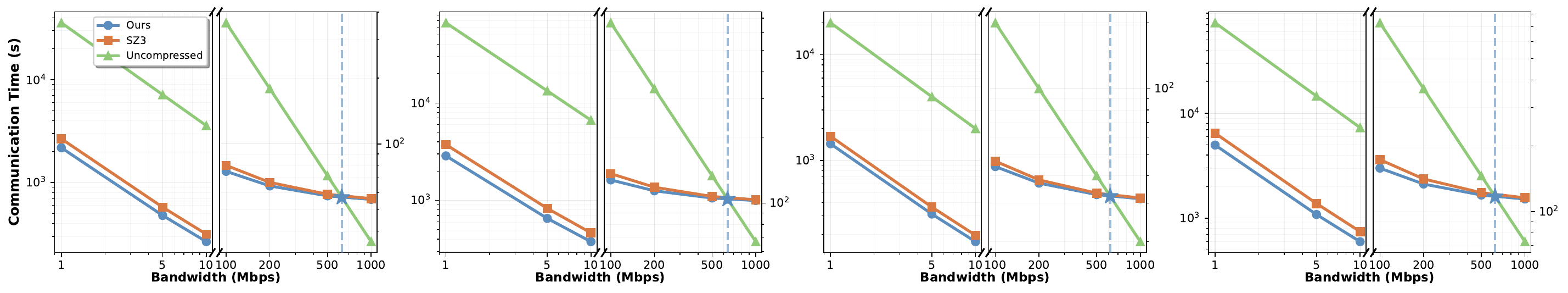}
        \label{fig:comm_time_over_bw}
    \end{subfigure}

    \vspace{-2em}
    
    \caption{End-to-end communication time. The upper panel compares various relative error bounds under the same network bandwidth (10 Mbps). The lower panel is for different network bandwidths under the same relative error bound (3e-2).  }
    \label{fig:comm_time_combined}
\end{figure*}

\subsection{Layer-Wise Performance Analyses}
When using the kernel-wise sign consistency strategy, only kernels whose sign consistency exceeds the predefined threshold are predicted, while the remaining kernels are compressed using the standard SZ3 pipeline without prediction. This means that all observed compression gains originate solely from the subset of kernels selected for prediction. To further examine this effect, we conduct a layer-wise analysis on the largest convolutional layer in ResNet-18, which contains $512\times512$ kernels of size $3\times3$. The experiment is performed on the CIFAR-10 dataset with a sign consistency threshold of 0.5 and a relative error bound of $3\times10^{-2}$. 

Figure~\ref{fig:conv layer comp analysis} illustrates the distributional changes and compression behavior within this layer.Figure~\ref{fig:conv layer comp analysis}(a) compares the data distribution of the kernels selected for prediction before and after applying our predictor. The blue curve represents the original kernel values, while the green curve shows the distribution of residuals after prediction. The residuals exhibit a much sharper concentration around zero, indicating reduced variance and entropy which directly contributing to better compressibility. Figure~\ref{fig:conv layer comp analysis}(b) presents the overall layer-wise distribution by merging the residuals of predicted kernels with the original values of unpredicted kernels. The orange curve represents the combined distribution, which still shows a clear central peak and thinner tails compared to the original blue distribution, confirming that prediction effectively densifies the representable range even at the layer level. Correspondingly, Figure~\ref{fig:conv layer comp analysis}(c) quantifies the compression ratios of different components. The predicted kernels originally achieved only 21.6$\times$ compression with SZ3, whereas our predictor increases this to 29.7$\times$. When combined with unpredicted kernels, the overall layer compression reaches 29.6$\times$, compared to 23.86$\times$ for the baseline SZ3 without prediction. This significant enhancement confirms that the introduced temporal and structural predictors substantially reduce redundancy in gradient data, resulting in more compact encoding within individual layers.

To further investigate the effect of kernel size on the layer-wise compression performance, we varied the convolutional kernel size from $3\times3$ to $5\times5$ and $7\times7$ under the same experimental setup described earlier. The results are summarized in Table~\ref{tab:kernel_analysis}. As a baseline, for the $3\times3$ kernels, 60.60\% of them exceed the sign consistency threshold of 0.5, indicating that the predictor is applied to this portion of kernels, while the sign prediction error rate remains as low as 9.70\%. The bitmap introduced to record prediction metadata contributes only 12.07\% overhead relative to the total compressed size, confirming its negligible impact. When increasing the kernel size to $5\times5$, all parts of the compression ratio show consistent improvement. The prediction ratio decreases only slightly, and the sign prediction error rate remains almost unchanged. Meanwhile, the relative bitmap overhead drops significantly to 7.95\%. This reduction occurs because the total number of kernels decreases while the overall data volume grows, making the bitmap a smaller fraction of the compressed output. Consequently, the overall compression benefit becomes more pronounced at this scale. However, as the kernel size further increases to $7\times7$, the trend changes. While the compression ratio of the predicted portion remains similar, the proportion of kernels eligible for prediction nearly halves, and the sign mismatch rate rises to 15.48\%. Although the bitmap overhead further declines to 3.43\%, the substantial drop in the number of predictable kernels offsets the potential gain. Moreover, lowering the consistency threshold to increase prediction coverage would simultaneously amplify the sign prediction error, degrading the residual compressibility and negating the overall benefit. Therefore, the compressor does not necessarily achieve better efficiency with larger kernel sizes. In practice, for commonly used $3\times3$ and $5\times5$ convolutional kernels, the proposed method consistently delivers reliable and significant compression improvements.

\subsection{End-to-End Communication Gains}
While the compression ratio provides a direct measure of data reduction, its ultimate impact in Federated Learning manifests as reduced end-to-end communication time. To evaluate this effect, we simulate constrained-bandwidth environments by calculating the expected transmission time under limited bandwidth and introducing artificial latency on top of the actual cluster throughput~\cite{wilkins2024fedsz}. The measured communication time refers to the total wall-clock duration from the end of a client’s local training to the point when the server successfully receives and reconstructs the model update. The total communication time consists of three parts, including local compression on the client, data transmission over a fixed-bandwidth network, and decompression on the server.

We measure the total communication time under various network bandwidths ranging from 1Mbps to 1Gbps and different relative error bounds. For each configuration, we simulate 100 training rounds of Federated Learning on CIFAR-10 across all models. Figure~11 summarizes the measured end-to-end communication time. The upper panel shows, for each model, the total communication time under a fixed 10~Mbps bandwidth and different error bounds, while the lower panel reports the variation of total communication time at a fixed relative error bound of $3\times10^{-2}$ across multiple bandwidths.

In the upper panel of Figure~\ref{fig:comm_time_combined}, the blue bars represent the total communication time achieved with our proposed compressor, the orange bars correspond to SZ3, and the green dashed line indicates the communication time without compression. Across all models and error bounds, our method consistently achieves shorter end-to-end latency, reducing total communication time by an average of 22.4\%. For instance, in Inception~V3 with a relative error bound of $3\times10^{-2}$, the total communication time decreases from 662s with SZ3 to 514s with our method. It is important to note that SZ3 has already achieved a substantial reduction in communication time. Compared to uncompressed transmission, our method further reduces the communication time by 76.1–96.2\%. Hence, our method achieves further improvements under already highly optimized conditions.

The lower panel of Figure~\ref{fig:comm_time_combined} further demonstrates that our method maintains consistent superiority across different network bandwidths. The left half of each subfigure corresponds to low-bandwidth scenarios (1–10 Mbps), while the right half shows high-bandwidth settings (100–1000 Mbps), and the overall trend remains stable across this range. As expected, the benefit of compression is most prominent under low-bandwidth constraints, where transmission dominates total latency. As bandwidth increases, the relative advantage gradually diminishes, since compression and decompression incur fixed overheads that become more noticeable when transmission time shortens. The bandwidth thresholds beyond which compression ceases to yield positive gains are marked with stars in the figure. For our method under the relative error bound of $3\times10^{-2}$, these thresholds typically appear around 620 Mbps. This indicates that in most realistic FL deployments where edge devices typically operate under significantly lower uplink speeds, the proposed method ensures substantial communication benefits. Moreover, the observed trend remains stable across models and does not strongly depend on model size. This is because both compression and decompression as well as transmission times scale proportionally with model parameters, while the attainable benefit primarily depends on the achieved compression ratio and the throughput of the compressor. Higher compression ratios and faster codec throughput can therefore further elevate the effective bandwidth threshold for positive gains.

Overall, these results confirm that the proposed predictive compression pipeline not only improves compression ratios but also translates these improvements into tangible end-to-end communication efficiency. The gain remains stable and predictable across different models, datasets, and network conditions, ensuring practical value in diverse real-world FL scenarios. These findings suggest that the proposed compressor can provide sustained performance advantages even under realistic communication constraints.

\section{Discussion and Future Work}
While our new EBLC demonstrates significant compression gains on gradient data across diverse models and datasets, there remain opportunities to further improve its adaptability and generality. One practical consideration is the presence of tunable hyperparameters in our design, such as the EMA decay factor, sign-consistency thresholds, and error bound. Although these parameters are relatively lightweight to configure, their optimal settings can vary across tasks and model architectures. In this work, we manually tune them based on accuracy and compression ratio. In future work, we aim to develop auto-tuning mechanisms that can dynamically adapt these parameters based on the observed gradient statistics during training. Such self-adaptive configurations would make the compressor more robust and deployable in heterogeneous FL environments. Moreover, the low algorithmic complexity and high parallelism of our predictor design make it well-suited for GPUs. We plan to develop an optimized GPU version that leverages modern parallel hardware to further improve throughput, enabling faster encoding and decoding in large-scale FL deployments.

\section{Related Works}
In this section, we discuss sparsification-based approaches and model compression techniques to clarify the positioning and uniqueness of our method.

\subsection{Sparsification-Based Compression}
Sparsification reduces communication overhead by selecting and transmitting only a subset of gradient elements, typically the top-$k$ or those above a threshold~\cite{aji2017sparse, mishchenko2024distributed, gorbunov2021marina, renggli2019sparcml,strom2015scalable}. Although effective in reducing volume, these methods introduce uncontrolled approximation error and rely on heuristic mechanisms like momentum correction or error feedback to maintain training stability~\cite{lin2017deep, karimireddy2019error,stich2018sparsified}. In contrast, our work belongs to the family of error-bounded lossy compressors (EBLCs), which explicitly constrain the deviation between original and reconstructed values, offering more predictable accuracy behavior during training. Moreover, our method is not mutually exclusive with sparsification. They target different forms of redundancy and can be composed sequentially. In fact, our predictor-enhanced EBLC can serve as a downstream quantizer applied to the selected subset in a sparsified gradient, further reducing transmission cost without violating error guarantees.

\subsection{Compression of Machine Learning Models}
A rich line of research has explored compressing deep neural networks to reduce storage, computation, and communication costs. Common approaches include model quantization~\cite{jacob2018quantization, banner2019post, hubara2018quantized}, which lowers parameter precision to fewer bits; weight pruning~\cite{han2015learning, gale2019state, he2023structured}, which removes redundant connections to yield sparse architectures; and low-rank decomposition~\cite{denton2014exploiting, xue2013restructuring, jaderberg2014speeding}, which approximates dense weight matrices using compact factorizations. These techniques effectively shrink model size before or during training and can also reduce the volume of data exchanged in each communication round of federated learning (FL). Moreover, mixed-precision training has been widely adopted to accelerate computation and alleviate memory traffic on modern hardware. Nevertheless, even with quantized or pruned models, the gradients exchanged between clients and server remain large. Our approach is fully orthogonal to these model compression techniques and can be applied on top of them to achieve additional communication savings. In addition, it introduces negligible runtime overhead and requires no architectural modification or retraining, making it practical for deployment in existing FL systems.

\section{Conclusion}
We present an error-bounded lossy compressor with a novel predictor tailored for gradient data in federated learning. By exploiting temporal patterns across training epochs and intra-kernel structural regularities, it produces substantially lower-entropy residuals, enabling higher compression ratios without sacrificing model accuracy compared to the highly optimized generic EBLC, SZ3. Integrated into the production-grade APPFL framework, our method delivers significant end-to-end communication efficiency improvements under bandwidth-constrained settings, demonstrating its practical scalability for real-world FL deployments.

\bibliographystyle{ACM-Reference-Format}
\bibliography{reference}

@inproceedings{sz2,
  title={Fast error-bounded lossy HPC data compression with SZ},
  author={Di, Sheng and Cappello, Franck},
  booktitle={2016 ieee international parallel and distributed processing symposium (ipdps)},
  pages={730--739},
  year={2016},
  organization={IEEE}
}

@inproceedings{sz3,
  title={Optimizing error-bounded lossy compression for scientific data by dynamic spline interpolation},
  author={Zhao, Kai and Di, Sheng and Dmitriev, Maxim and Tonellot, Thierry-Laurent D and Chen, Zizhong and Cappello, Franck},
  booktitle={2021 IEEE 37th International Conference on Data Engineering (ICDE)},
  pages={1643--1654},
  year={2021},
  organization={IEEE}
}

@article{SZ3_BD,
  title={Sz3: A modular framework for composing prediction-based error-bounded lossy compressors},
  author={Liang, Xin and Zhao, Kai and Di, Sheng and Li, Sihuan and Underwood, Robert and Gok, Ali M and Tian, Jiannan and Deng, Junjing and Calhoun, Jon C and Tao, Dingwen and others},
  journal={IEEE Transactions on Big Data},
  volume={9},
  number={2},
  pages={485--498},
  year={2022},
  publisher={IEEE}
}

@inproceedings{grad_oscillation,
  title={Exploring Gradient Oscillation in Deep Neural Network Training},
  author={Morchdi, Chedi and Zhou, Yi and Ding, Jie and Wang, Bei},
  booktitle={2023 59th Annual Allerton Conference on Communication, Control, and Computing (Allerton)},
  pages={1--7},
  year={2023},
  organization={IEEE}
}

@article{aji2017sparse,
  title={Sparse communication for distributed gradient descent},
  author={Aji, Alham Fikri and Heafield, Kenneth},
  journal={arXiv preprint arXiv:1704.05021},
  year={2017}
}

@article{lin2017deep,
  title={Deep gradient compression: Reducing the communication bandwidth for distributed training},
  author={Lin, Yujun and Han, Song and Mao, Huizi and Wang, Yu and Dally, William J},
  journal={arXiv preprint arXiv:1712.01887},
  year={2017}
}

@inproceedings{karimireddy2019error,
  title={Error feedback fixes signsgd and other gradient compression schemes},
  author={Karimireddy, Sai Praneeth and Rebjock, Quentin and Stich, Sebastian and Jaggi, Martin},
  booktitle={International conference on machine learning},
  pages={3252--3261},
  year={2019},
  organization={PMLR}
}

@inproceedings{jacob2018quantization,
  title={Quantization and training of neural networks for efficient integer-arithmetic-only inference},
  author={Jacob, Benoit and Kligys, Skirmantas and Chen, Bo and Zhu, Menglong and Tang, Matthew and Howard, Andrew and Adam, Hartwig and Kalenichenko, Dmitry},
  booktitle={Proceedings of the IEEE conference on computer vision and pattern recognition},
  pages={2704--2713},
  year={2018}
}

@article{banner2019post,
  title={Post training 4-bit quantization of convolutional networks for rapid-deployment},
  author={Banner, Ron and Nahshan, Yury and Soudry, Daniel},
  journal={Advances in neural information processing systems},
  volume={32},
  year={2019}
}

@article{han2015learning,
  title={Learning both weights and connections for efficient neural network},
  author={Han, Song and Pool, Jeff and Tran, John and Dally, William},
  journal={Advances in neural information processing systems},
  volume={28},
  year={2015}
}

@article{gale2019state,
  title={The state of sparsity in deep neural networks},
  author={Gale, Trevor and Elsen, Erich and Hooker, Sara},
  journal={arXiv preprint arXiv:1902.09574},
  year={2019}
}

@article{denton2014exploiting,
  title={Exploiting linear structure within convolutional networks for efficient evaluation},
  author={Denton, Emily L and Zaremba, Wojciech and Bruna, Joan and LeCun, Yann and Fergus, Rob},
  journal={Advances in neural information processing systems},
  volume={27},
  year={2014}
}

@inproceedings{xue2013restructuring,
  title={Restructuring of deep neural network acoustic models with singular value decomposition.},
  author={Xue, Jian and Li, Jinyu and Gong, Yifan},
  booktitle={Interspeech},
  pages={2365--2369},
  year={2013}
}

@article{yang2022flash,
  title={Flash: Heterogeneity-aware federated learning at scale},
  author={Yang, Chengxu and Xu, Mengwei and Wang, Qipeng and Chen, Zhenpeng and Huang, Kang and Ma, Yun and Bian, Kaigui and Huang, Gang and Liu, Yunxin and Jin, Xin and others},
  journal={IEEE Transactions on Mobile Computing},
  volume={23},
  number={1},
  pages={483--500},
  year={2022},
  publisher={IEEE}
}

@article{bonawitz2019towards,
  title={Towards federated learning at scale: System design},
  author={Bonawitz, Keith and Eichner, Hubert and Grieskamp, Wolfgang and Huba, Dzmitry and Ingerman, Alex and Ivanov, Vladimir and Kiddon, Chloe and Kone{\v{c}}n{\`y}, Jakub and Mazzocchi, Stefano and McMahan, Brendan and others},
  journal={Proceedings of machine learning and systems},
  volume={1},
  pages={374--388},
  year={2019}
}

@article{gabrielli2023survey,
  title={A survey on decentralized federated learning},
  author={Gabrielli, Edoardo and Pica, Giovanni and Tolomei, Gabriele},
  journal={arXiv preprint arXiv:2308.04604},
  year={2023}
}

@inproceedings{guo2021multi,
  title={Multi-institutional collaborations for improving deep learning-based magnetic resonance image reconstruction using federated learning},
  author={Guo, Pengfei and Wang, Puyang and Zhou, Jinyuan and Jiang, Shanshan and Patel, Vishal M},
  booktitle={Proceedings of the IEEE/CVF conference on computer vision and pattern recognition},
  pages={2423--2432},
  year={2021}
}

@article{darestani2025convolutional,
  title={Convolutional neural network based system for fully automatic FLAIR MRI segmentation in multiple sclerosis diagnosis},
  author={Darestani, Ali Arian and Davarani, Mahsa Naeeni and -Ca{\~n}as, Virginia Guillen and Hashemi, Hasan and Zarei, Amin and Havadaragh, Sanaz Heydari and Harirchian, Mohammad Hossein},
  journal={Scientific Reports},
  volume={15},
  number={1},
  pages={36146},
  year={2025},
  publisher={Nature Publishing Group UK London}
}

@article{kairouz2021advances,
  title={Advances and open problems in federated learning},
  author={Kairouz, Peter and McMahan, H Brendan and Avent, Brendan and Bellet, Aur{\'e}lien and Bennis, Mehdi and Bhagoji, Arjun Nitin and Bonawitz, Kallista and Charles, Zachary and Cormode, Graham and Cummings, Rachel and others},
  journal={Foundations and trends{\textregistered} in machine learning},
  volume={14},
  number={1--2},
  pages={1--210},
  year={2021},
  publisher={Now Publishers, Inc.}
}

@article{alistarh2017qsgd,
  title={QSGD: Communication-efficient SGD via gradient quantization and encoding},
  author={Alistarh, Dan and Grubic, Demjan and Li, Jerry and Tomioka, Ryota and Vojnovic, Milan},
  journal={Advances in neural information processing systems},
  volume={30},
  year={2017}
}

@inproceedings{ryu2022appfl,
  title={APPFL: open-source software framework for privacy-preserving federated learning},
  author={Ryu, Minseok and Kim, Youngdae and Kim, Kibaek and Madduri, Ravi K},
  booktitle={2022 IEEE international parallel and distributed processing symposium workshops (IPDPSW)},
  pages={1074--1083},
  year={2022},
  organization={IEEE}
}

@article{wang2019adaptive,
  title={Adaptive federated learning in resource constrained edge computing systems},
  author={Wang, Shiqiang and Tuor, Tiffany and Salonidis, Theodoros and Leung, Kin K and Makaya, Christian and He, Ting and Chan, Kevin},
  journal={IEEE journal on selected areas in communications},
  volume={37},
  number={6},
  pages={1205--1221},
  year={2019},
  publisher={IEEE}
}

@inproceedings{mcmahan2017communication,
  title={Communication-efficient learning of deep networks from decentralized data},
  author={McMahan, Brendan and Moore, Eider and Ramage, Daniel and Hampson, Seth and y Arcas, Blaise Aguera},
  booktitle={Artificial intelligence and statistics},
  pages={1273--1282},
  year={2017},
  organization={PMLR}
}

@article{dalcin2021mpi4py,
  title={mpi4py: Status update after 12 years of development},
  author={Dalcin, Lisandro and Fang, Yao-Lung L},
  journal={Computing in Science \& Engineering},
  volume={23},
  number={4},
  pages={47--54},
  year={2021},
  publisher={IEEE}
}

@inproceedings{he2016deep,
  title={Deep residual learning for image recognition},
  author={He, Kaiming and Zhang, Xiangyu and Ren, Shaoqing and Sun, Jian},
  booktitle={Proceedings of the IEEE conference on computer vision and pattern recognition},
  pages={770--778},
  year={2016}
}

@inproceedings{szegedy2015going,
  title={Going deeper with convolutions},
  author={Szegedy, Christian and Liu, Wei and Jia, Yangqing and Sermanet, Pierre and Reed, Scott and Anguelov, Dragomir and Erhan, Dumitru and Vanhoucke, Vincent and Rabinovich, Andrew},
  booktitle={Proceedings of the IEEE conference on computer vision and pattern recognition},
  pages={1--9},
  year={2015}
}

@inproceedings{szegedy2016rethinking,
  title={Rethinking the inception architecture for computer vision},
  author={Szegedy, Christian and Vanhoucke, Vincent and Ioffe, Sergey and Shlens, Jon and Wojna, Zbigniew},
  booktitle={Proceedings of the IEEE conference on computer vision and pattern recognition},
  pages={2818--2826},
  year={2016}
}

@article{xiao2017fashion,
  title={Fashion-mnist: a novel image dataset for benchmarking machine learning algorithms},
  author={Xiao, Han and Rasul, Kashif and Vollgraf, Roland},
  journal={arXiv preprint arXiv:1708.07747},
  year={2017}
}

@article{krizhevsky2009learning,
  title={Learning multiple layers of features from tiny images},
  author={Krizhevsky, Alex and Hinton, Geoffrey and others},
  year={2009},
  publisher={Toronto, ON, Canada}
}

@inproceedings{fei2004learning,
  title={Learning generative visual models from few training examples: An incremental bayesian approach tested on 101 object categories},
  author={Fei-Fei, Li and Fergus, Rob and Perona, Pietro},
  booktitle={2004 conference on computer vision and pattern recognition workshop},
  pages={178--178},
  year={2004},
  organization={IEEE}
}

@article{kalman1960new,
  title={A new approach to linear filtering and prediction problems},
  author={Kalman, Rudolph Emil},
  year={1960}
}

@inproceedings{zhou2021informer,
  title={Informer: Beyond efficient transformer for long sequence time-series forecasting},
  author={Zhou, Haoyi and Zhang, Shanghang and Peng, Jieqi and Zhang, Shuai and Li, Jianxin and Xiong, Hui and Zhang, Wancai},
  booktitle={Proceedings of the AAAI conference on artificial intelligence},
  volume={35},
  number={12},
  pages={11106--11115},
  year={2021}
}

@article{liu2023itransformer,
  title={itransformer: Inverted transformers are effective for time series forecasting},
  author={Liu, Yong and Hu, Tengge and Zhang, Haoran and Wu, Haixu and Wang, Shiyu and Ma, Lintao and Long, Mingsheng},
  journal={arXiv preprint arXiv:2310.06625},
  year={2023}
}

@article{mishchenko2024distributed,
  title={Distributed learning with compressed gradient differences},
  author={Mishchenko, Konstantin and Gorbunov, Eduard and Tak{\'a}{\v{c}}, Martin and Richt{\'a}rik, Peter},
  journal={Optimization Methods and Software},
  pages={1--16},
  year={2024},
  publisher={Taylor \& Francis}
}

@inproceedings{gorbunov2021marina,
  title={MARINA: Faster non-convex distributed learning with compression},
  author={Gorbunov, Eduard and Burlachenko, Konstantin P and Li, Zhize and Richt{\'a}rik, Peter},
  booktitle={International Conference on Machine Learning},
  pages={3788--3798},
  year={2021},
  organization={PMLR}
}

@inproceedings{wilkins2024fedsz,
  title={Fedsz: Leveraging error-bounded lossy compression for federated learning communications},
  author={Wilkins, Grant and Di, Sheng and Calhoun, Jon C and Li, Zilinghan and Kim, Kibaek and Underwood, Robert and Mortier, Richard and Cappello, Franck},
  booktitle={2024 IEEE 44th International Conference on Distributed Computing Systems (ICDCS)},
  pages={577--588},
  year={2024},
  organization={IEEE}
}

@article{lindstrom2014fixed,
  title={Fixed-rate compressed floating-point arrays},
  author={Lindstrom, Peter},
  journal={IEEE transactions on visualization and computer graphics},
  volume={20},
  number={12},
  pages={2674--2683},
  year={2014},
  publisher={IEEE}
}

@article{ainsworth2019multilevel,
  title={Multilevel techniques for compression and reduction of scientific data---the multivariate case},
  author={Ainsworth, Mark and Tugluk, Ozan and Whitney, Ben and Klasky, Scott},
  journal={SIAM Journal on Scientific Computing},
  volume={41},
  number={2},
  pages={A1278--A1303},
  year={2019},
  publisher={SIAM}
}

@article{rieke2020future,
  title={The future of digital health with federated learning},
  author={Rieke, Nicola and Hancox, Jonny and Li, Wenqi and Milletari, Fausto and Roth, Holger R and Albarqouni, Shadi and Bakas, Spyridon and Galtier, Mathieu N and Landman, Bennett A and Maier-Hein, Klaus and others},
  journal={NPJ digital medicine},
  volume={3},
  number={1},
  pages={119},
  year={2020},
  publisher={Nature Publishing Group UK London}
}

@article{pokhrel2020federated,
  title={Federated learning with blockchain for autonomous vehicles: Analysis and design challenges},
  author={Pokhrel, Shiva Raj and Choi, Jinho},
  journal={IEEE Transactions on Communications},
  volume={68},
  number={8},
  pages={4734--4746},
  year={2020},
  publisher={IEEE}
}

@article{stich2018sparsified,
  title={Sparsified SGD with memory},
  author={Stich, Sebastian U and Cordonnier, Jean-Baptiste and Jaggi, Martin},
  journal={Advances in neural information processing systems},
  volume={31},
  year={2018}
}

@article{hubara2018quantized,
  title={Quantized neural networks: Training neural networks with low precision weights and activations},
  author={Hubara, Itay and Courbariaux, Matthieu and Soudry, Daniel and El-Yaniv, Ran and Bengio, Yoshua},
  journal={journal of machine learning research},
  volume={18},
  number={187},
  pages={1--30},
  year={2018}
}

@article{jaderberg2014speeding,
  title={Speeding up convolutional neural networks with low rank expansions},
  author={Jaderberg, Max and Vedaldi, Andrea and Zisserman, Andrew},
  journal={arXiv preprint arXiv:1405.3866},
  year={2014}
}

@article{he2023structured,
  title={Structured pruning for deep convolutional neural networks: A survey},
  author={He, Yang and Xiao, Lingao},
  journal={IEEE transactions on pattern analysis and machine intelligence},
  volume={46},
  number={5},
  pages={2900--2919},
  year={2023},
  publisher={IEEE}
}

@inproceedings{huang2023cuszp,
  title={cuszp: An ultra-fast gpu error-bounded lossy compression framework with optimized end-to-end performance},
  author={Huang, Yafan and Di, Sheng and Yu, Xiaodong and Li, Guanpeng and Cappello, Franck},
  booktitle={Proceedings of the International Conference for High Performance Computing, Networking, Storage and Analysis},
  pages={1--13},
  year={2023}
}

@inproceedings{renggli2019sparcml,
  title={SparCML: High-performance sparse communication for machine learning},
  author={Renggli, C{\`e}dric and Ashkboos, Saleh and Aghagolzadeh, Mehdi and Alistarh, Dan and Hoefler, Torsten},
  booktitle={Proceedings of the International Conference for High Performance Computing, Networking, Storage and Analysis},
  pages={1--15},
  year={2019}
}

@article{wu2021autoformer,
  title={Autoformer: Decomposition transformers with auto-correlation for long-term series forecasting},
  author={Wu, Haixu and Xu, Jiehui and Wang, Jianmin and Long, Mingsheng},
  journal={Advances in neural information processing systems},
  volume={34},
  pages={22419--22430},
  year={2021}
}

@inproceedings{tao2017significantly,
  title={Significantly improving lossy compression for scientific data sets based on multidimensional prediction and error-controlled quantization},
  author={Tao, Dingwen and Di, Sheng and Chen, Zizhong and Cappello, Franck},
  booktitle={2017 IEEE International Parallel and Distributed Processing Symposium (IPDPS)},
  pages={1129--1139},
  year={2017},
  organization={IEEE}
}

@article{strom2015scalable,
  title={Scalable distributed DNN training using commodity GPU cloud computing},
  author={Str{\"o}m, Nikko},
  year={2015}
}

@article{zhang2025fedcspc,
  title={FedCSpc: A Cross-Silo Federated Learning System with Error-Bounded Lossy Parameter Compression},
  author={Zhang, Zhaorui and Di, Sheng and Zhao, Kai and Jin, Sian and Tao, Dingwen and Ji, Zhuoran and Liu, Benben and Alharthi, Khalid Ayed and Cao, Jiannong and Cappello, Franck},
  journal={IEEE Transactions on Parallel and Distributed Systems},
  year={2025},
  publisher={IEEE}
}

@article{zhang2025fedefsz,
  title={FedEFsz: Fair Cross-Silo Federated Learning System with Error-Bounded Lossy Compression},
  author={Zhang, Zhaorui and Di, Sheng and Liu, Benben and Ji, Zhuoran and Li, Guanpeng and Lu, Xiaoyi and Zhou, Amelie Chi and Alharthi, Khalid Ayed and Cao, Jiannong},
  journal={IEEE Transactions on Parallel and Distributed Systems},
  year={2025},
  publisher={IEEE}
}

\end{document}